\documentclass{bmvc2k}
\usepackage{tikz}
\def\checkmark{\tikz\fill[scale=0.25](0,.35) -- (.25,0) -- (1,.7) -- (.25,.15) -- cycle;} 
\usepackage{paralist, tabularx}
\usepackage[ruled]{algorithm2e}
\usepackage{enumitem}
\usepackage{multirow}
\usepackage{amssymb}
\makeatletter
\def\BState{\State\hskip-\ALG@thistlm}
\makeatother

\title{\hspace{0.4cm}Attend Refine Repeat: Active Box Proposal\\\hspace{0.4cm}Generation via In-Out Localization}

\addauthor{Spyros Gidaris}{spyros.gidaris@enpc.fr}{1}
\addauthor{Nikos Komodakis}{nikos.komodakis@enpc.fr}{1}

\addinstitution{
 Universit\'e Paris-Est, \'Ecole des Ponts ParisTech\\
 Paris, France
}
\runninghead{S. Gidaris and N. Komodakis}{Attend Refine Repeat}

\def\etal{\emph{et al}\bmvaOneDot}

\begin{document}

\maketitle

\begin{abstract}
The problem of computing category agnostic bounding box proposals is utilized as a core component in many computer vision tasks and thus has lately attracted a lot of attention.
In this work we propose a new approach to tackle this problem that is based on an active strategy for generating box proposals that starts from a set of seed boxes, 
which are uniformly distributed on the image, and then progressively moves its attention on the promising image areas where it is more likely to discover well localized bounding box proposals. 
We call our approach \emph{AttractioNet} and a core component of it is a CNN-based category agnostic object location refinement module that is capable of yielding accurate and robust bounding box predictions regardless of the object category.
 
We extensively evaluate our \emph{AttractioNet} approach on several image datasets (i.e. COCO, PASCAL, ImageNet detection and NYU-Depth V2 datasets) reporting on all of them state-of-the-art results that surpass the previous work in the field by a significant margin and also providing strong empirical evidence that our approach is capable to generalize to unseen categories.
Furthermore, 
we evaluate our \emph{AttractioNet} proposals in the context of the object detection task using a VGG16-Net based detector and the achieved detection performance on COCO manages to significantly surpass all other VGG16-Net based detectors while even being competitive with a heavily tuned ResNet-101 based detector.  Code as well as box proposals computed for several datasets are available at:: \url{https://github.com/gidariss/AttractioNet}.
\end{abstract}

\section{Introduction}
\label{sec:intro}

Category agnostic object proposal generation is a computer vision task that has received an immense amount of attention over the last years.
Its definition is that for a given image a small set of instance segmentations or bounding boxes must be generated that will cover with high recall all the objects that appear in the image regardless of their category. 
In object detection, applying the recognition models to such a reduced set of category independent location hypothesis~\cite{girshick2014rich} instead of an exhaustive scan of the entire image~\cite{felzenszwalb2010object,sermanet2013overfeat}, has the advantages of drastically reducing the amount of recognition model evaluations and thus allowing the use of more sophisticated machinery for that purpose.
As a result, proposal based detection systems manage to achieve state-of-the-art results and have become the dominant paradigm in the object detection literature~\cite{girshick2014rich,he2015spatial,girshick2015fast,gidaris2015object,gidaris2016locnet,shaoqing2015faster,zagoruyko2016multipath,bell2015inside,shrivastava2016training}.
Object proposals have also been used in various other tasks, such as weakly-supervised object detection~\cite{cinbis2015weakly}, exemplar 2D-3D detection~\cite{MassaCVPR16}, visual semantic role labelling~\cite{gupta2015visual}, caption generation~\cite{karpathy2015deep} or visual question answering~\cite{shih2015look}.

In this work we  focus on the problem of generating bounding box object proposals rather than instance segmentations. 
Several approaches have been proposed in the literature for this task~\cite{van2011segmentation,APBMM2014,zitnick2014edge,krahenbuhllearning,krahenbuhl2014geodesic,alexe2012measuring,LuJL15,cheng2014bing,hayder2016learning,chen2015improving,DaiHLRS16}.
Among them our work is most related to the CNN-based objectness scoring approaches~\cite{kuo2015deepbox,ghodrati2015deepproposal,pinheiro2015learning} that recently have demonstrated state-of-the-art results~\cite{pinheiro2015learning,PinheiroLCD16}.

In the objectness scoring paradigm, a large set of image boxes is ranked according to how likely it is for each image box to tightly enclose an object --- regardless of its category --- and then this set is post-processed with a non-maximum-suppression step and truncated to yield the final set of object proposals.
In this context, Kuo \etal~\cite{kuo2015deepbox} with their DeepBox system demonstrated that training a convolutional neural network to perform the task of objectness scoring can yield superior performance over previous methods that were based on low level cues and they provided empirical evidence that it can generalize to unseen categories.
In order to avoid evaluating the computationally expensive CNN-based objectness scoring model on hundreds of thousands image boxes,
which is necessary for achieving good localization of all the objects in the image,
they use it only to re-rank the proposals generated from a faster but less accurate proposal generator 
thus being limited by its localization performance.
Instead, more recent CNN-based approaches apply their models only to ten of thousands image boxes, uniformly distributed in the image, and jointly with objectness prediction they also infer the bounding box of the closest object to each input image box. 
Specifically, the Region Proposal Network in Faster-RCNN~\cite{shaoqing2015faster} performs bounding box regression for that purpose while the DeepMask method predicts the foreground mask of the object centred in the image box and then it infers the location of the object's bounding box by extracting the box that tightly encloses the foreground pixels. The latter has demonstrated state-of-the-art results and was recently extended with a top-down foreground mask refinement mechanism that exploits the convolutional feature maps at multiple depths of a neural network~\cite{PinheiroLCD16}. 

Our work is also based on the paradigm of having a CNN model that given an image box it jointly predicts its objectness and a new bounding box that is better aligned on the object that it contains. However, we opt to advance the previous state-of-the-art in box proposal generation in two ways: 
\textbf{(1)} improving the object's bounding box prediction step
\textbf{(2)} actively generating the set of image boxes that will be processed by the CNN model.

Regarding the bounding box inference step we exploit the recent advances in object detection where Gidaris and Komodakis~\cite{gidaris2016locnet} showed how to improve the object-specific localization accuracy.
Specifically, they replaced the bounding box regression step with a localization module, called \emph{LocNet}, that given a search region it infers the bounding box of the object inside the search region by assigning membership probabilities to each row and each column of that region and they empirically proved that this localization task is easier to be learned from a convolutional neural network thus yielding more accurate box predictions during test time. 
Given the importance of having accurate bounding box locations in the proposal generation task, 
we believe that it would be of great interest to develop and study a \textit{category agnostic} version of LocNet for this task.

Our second idea for improving the box proposal generation task stems from the following observation. 
Recent state-of-the-art box proposal methods evaluate only 
a relatively small set of image boxes (in the order of $10k$) uniformly distributed in the image and rely on the bounding box prediction step to fix the localization errors.
However, depending on how far  an object is from the closest evaluated image box, both the objectness scoring and the bounding box prediction for that object could be imperfect.
For instance, Hosang \etal~\cite{hosang2015makes} showed that in the case of the detection task the correct recognition of an object from an image box is correlated with how well the box encloses the object.
Given how similar are the tasks of category-specific object detection and category-agnostic proposal generation, it is safe to assume that a similar behaviour will probably hold for the latter one as well.
Hence, in our work we opt for an \textit{active} object localization scheme, which we call \emph{Attend Refine Repeat} algorithm, that starting from a set of seed boxes it progressively generates newer boxes that are expected with higher probability to be on the neighbourhood or to tightly enclose the objects of the image. 
Thanks to this localization scheme, our box proposal system is capable to both correct initially imperfect bounding box predictions and to give higher objectness score to candidate boxes that are more well localized on the objects of the image. 

To summarize, our contributions with respect to the box proposal generation task are: 
\begin{compactitem}
\item 
We developed a box proposal system that is based on an improved category-agnostic object location refinement module and on an active box proposal generation strategy that behaves as an attention mechanism that focus on the promising image areas in order to propose objects. 
We call the developed box proposal system \emph{AttractioNet}: \emph{(Att)end (R)efine Repeat: (Act)ive Box Proposal Generation via (I)n-(O)ut Localization (Net)work}.
\item
We exhaustively evaluate our system both on PASCAL and on the more challenging COCO datasets and we demonstrate significant improvement with respect to the state-of-the-art on box proposal generation. 
Furthermore, we provide strong evidence that our object location refinement module is capable of generalizing to unseen categories 
by reporting results for the unseen categories of ImageNet detection task and NYU-Depth dataset. 
\item 
Finally, we evaluate our box proposal generation approach in the context of the object detection task using a VGG16-Net based detection system 
and the achieved average precision performance on the COCO test-dev set manages to significantly surpass all  other VGG16-Net based detection systems while even being on par with the ResNet-101 based detection system of He \etal~\cite{he2015deep}.
\end{compactitem}

The remainder of the paper is structured as follows:
We describe our box proposal methodology in section \S2, 
we show experimental results in section \S3 and we  present our conclusions in section \S4.
\vspace{-10pt}
\section{Our approach} \label{sec:approach}
\subsection{Active bounding box proposal generation} \label{sec:algorithm}

\vspace{-10pt}
\begin{algorithm}\label{algo:prop_gen}
\renewcommand{\thealgocf}{}
\SetKwInOut{Input}{Input}
\SetKwInOut{Output}{Output}
\Input{Image $\textbf{I}$}
\Output{Bounding box proposals $\textbf{P}$}
$\textbf{C}  \gets \emptyset \text{, } \textbf{B}^{0} \gets \text{seed boxes}$\\
\For{$t \gets 1$ \textbf{to} $T$} {
$\texttt{/* Attend \& Refine procedure */}$\\
$\textbf{O}^{t} \gets \textit{ObjectnessScoring}(\textbf{B}^{t-1}| \textbf{I})$ \\
$\textbf{B}^{t} \gets \textit{ObjectLocationRefinement}(\textbf{B}^{t-1}| \textbf{I})$ \\
$\textbf{C}  \gets \textbf{C} \cup \{\textbf{B}^{t},\textbf{O}^{t}\}$\\ 
}
$\textbf{P} \gets \textit{NonMaxSuppression}(\textbf{C})$\\  
\caption{\textbf{\emph{Attend Refine Repeat}}}
\end{algorithm}

\begin{figure}
\begin{center}
\bmvaHangBox{
\includegraphics[width=2.5cm]{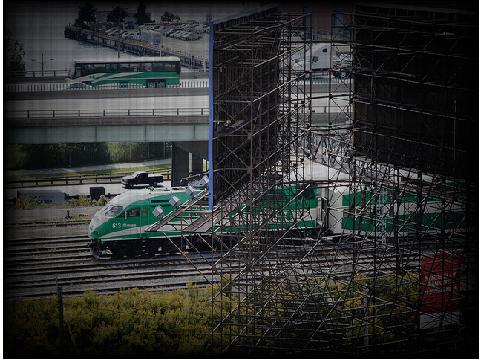}
\includegraphics[width=2.5cm]{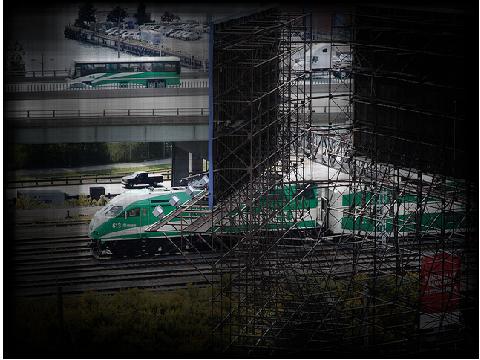}
\includegraphics[width=2.5cm]{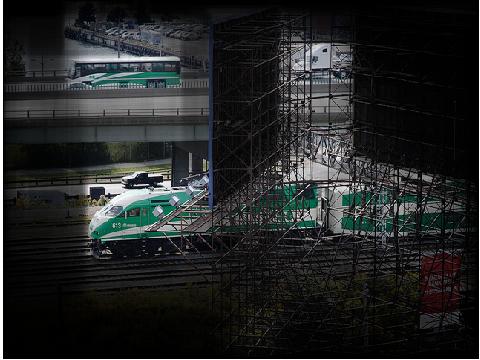}
\includegraphics[width=2.5cm]{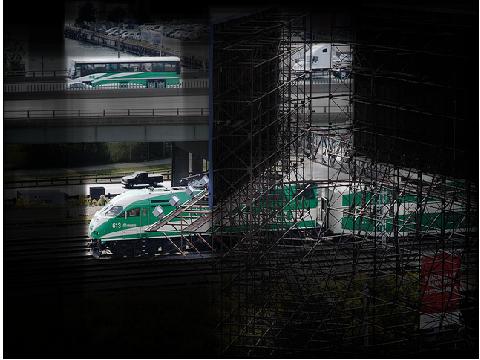}
\includegraphics[width=2.5cm]{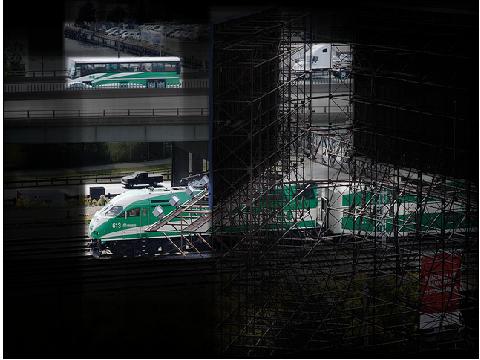}}\\
\vspace{-3pt}
\bmvaHangBox{
\includegraphics[width=2.5cm]{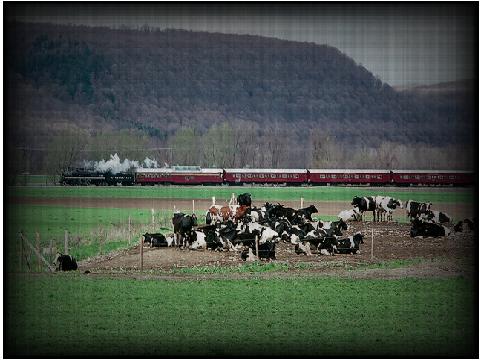}
\includegraphics[width=2.5cm]{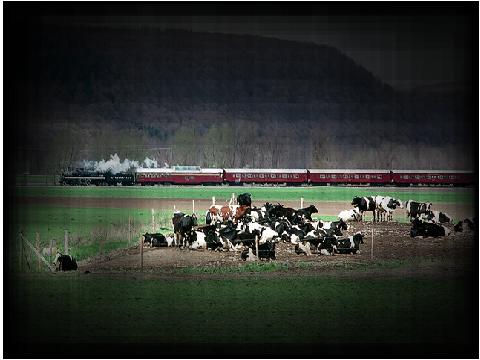}
\includegraphics[width=2.5cm]{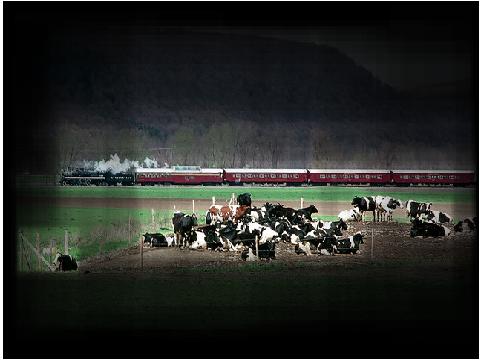}
\includegraphics[width=2.5cm]{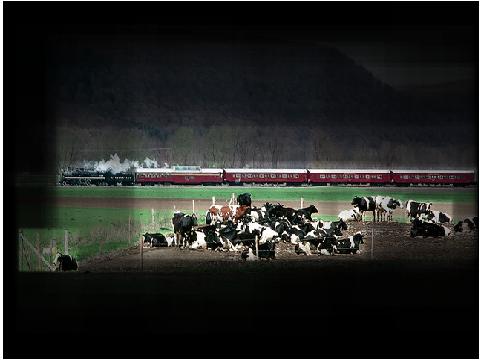}
\includegraphics[width=2.5cm]{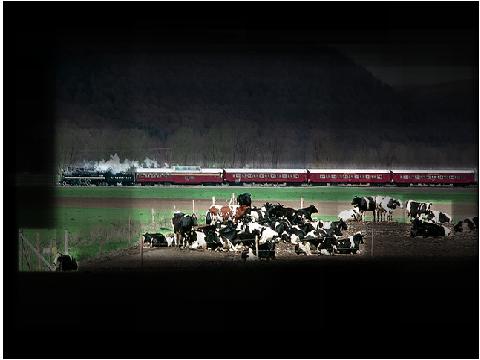}}\\
\vspace{-3pt}
\bmvaHangBox{
\includegraphics[width=2.5cm]{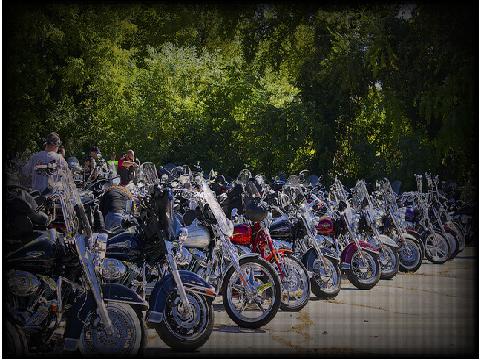}
\includegraphics[width=2.5cm]{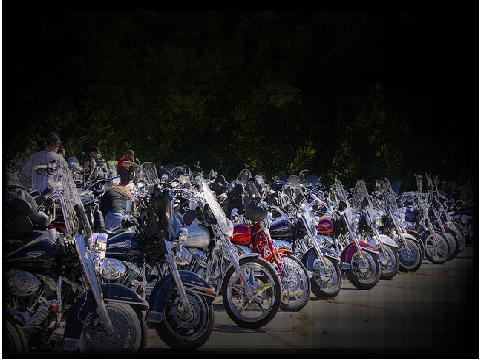}
\includegraphics[width=2.5cm]{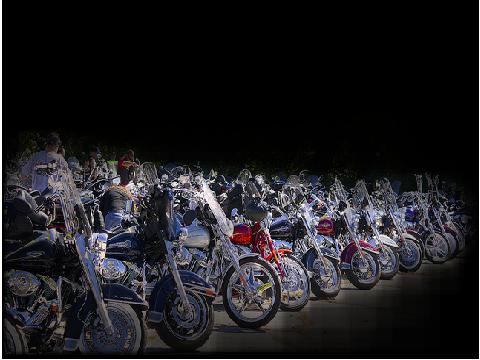}
\includegraphics[width=2.5cm]{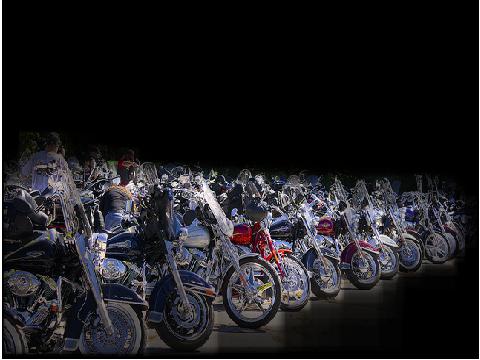}
\includegraphics[width=2.5cm]{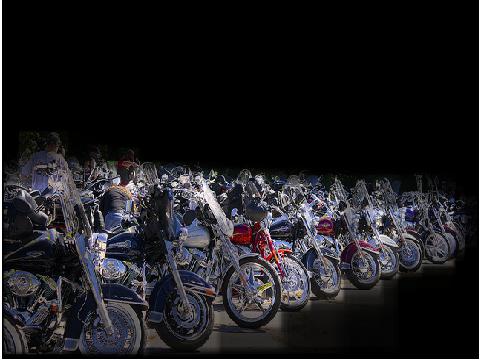}}\\
\vspace{-3pt}
\bmvaHangBox{
\includegraphics[width=2.5cm]{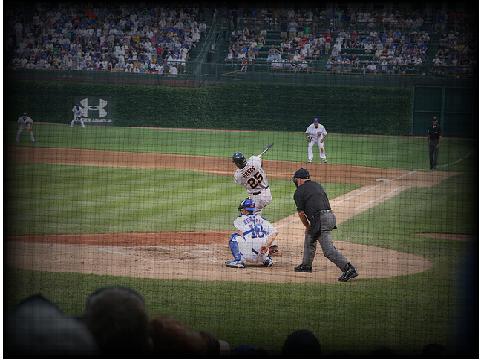}
\includegraphics[width=2.5cm]{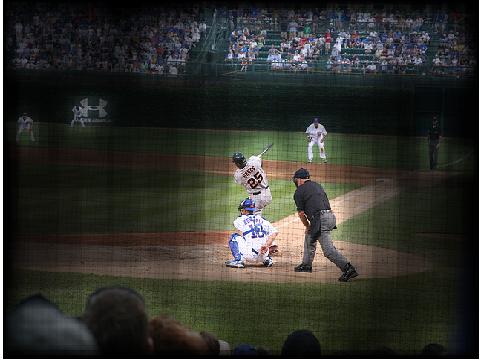}
\includegraphics[width=2.5cm]{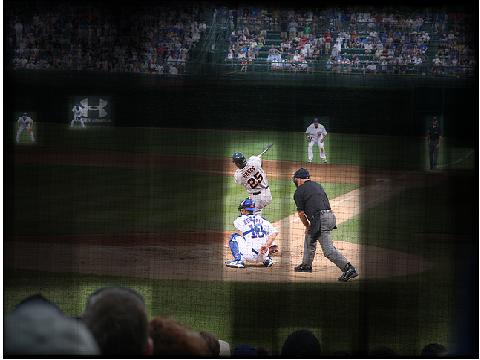}
\includegraphics[width=2.5cm]{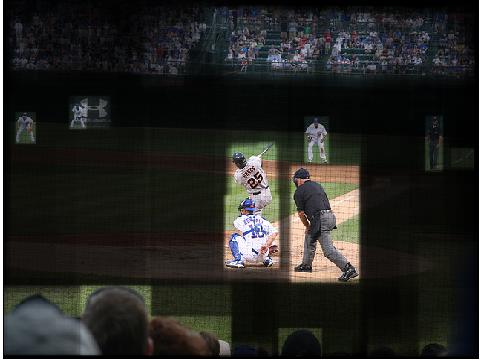}
\includegraphics[width=2.5cm]{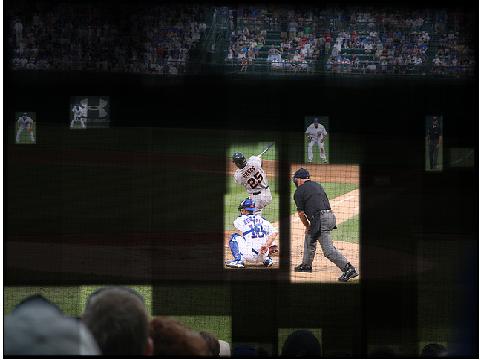}}\\
\vspace{-10pt}
\end{center}
\caption{\small{Illustration of the image areas  being attended by our box proposal generator algorithm at each iteration. In the first iteration the box proposal generator attends  the entire image since the seed boxes are created by uniformly distributing boxes across the image.
However, as the algorithm progresses its attention is concentrated on the image areas that actually contain objects.}}
\label{fig:Attentionmaps}
\vspace{-12pt}
\end{figure}

The active box proposal generation strategy that we employ in our work, which we call \emph{Attend Refine Repeat} algorithm, 
starts from a set of seed boxes, which only depend on the image size, and it then sequentially produces newer boxes that will better cover the objects of the image while avoiding the "objectless" image areas (see Figure~\ref{fig:Attentionmaps}). 
At the core of this algorithm lies a CNN-based box proposal model that, given an image $I$ and the coordinates of a box $B$,  executes the following operations:
\begin{compactdesc}
\item[Category agnostic object location refinement:] this operation returns the coordinates of a new box $\tilde{B}$ 
that would be more tightly aligned on the object near $B$. In case there are more than one objects in the neighbourhood of $B$ then the new box $\tilde{B}$  should be targeting  the object closest to the input box $B$, where by closest we mean the object that its bounding box has the highest intersection over union (IoU) overlap with the input box $B$.
\item[Category agnostic objectness scoring:] this operation scores the box $B$ based on how likely it is to tightly enclose an object, regardless of its category.
\end{compactdesc}
The pseudo-code of the \emph{Attend Refine Repeat} algorithm is provided in Algorithm~\ref{algo:prop_gen}. 
Specifically, it starts by initializing the set of candidate boxes $\textbf{C}$ to the empty set and then creates a set of seed boxes $\textbf{B}^{0}$ by uniformly distributing boxes of various fixed sizes in the image (similar to Cracking Bing~\cite{zhao2014cracking}). 
Then on each iteration $t$ it estimates the objectness $\textbf{O}^{t}$ of the boxes generated in the previous iteration, $\textbf{B}^{t-1}$, and it refines their location (resulting in boxes $\textbf{B}^{t}$) by attempting to predict the bounding boxes of the objects that are closest to them. 
The results $\{\textbf{B}^{t},\textbf{O}^{t}\}$ of those operations are added to the candidates set $\textbf{C}$ and the algorithm continues. In the end, non-maximum-suppression~\cite{felzenszwalb2010object} is applied to the candidate box proposals $\textbf{C}$ and the top $K$ box proposals, set $\textbf{P}$, are returned.

\begin{figure}
\begin{center}
\bmvaHangBox{
\includegraphics[width=2.0cm]{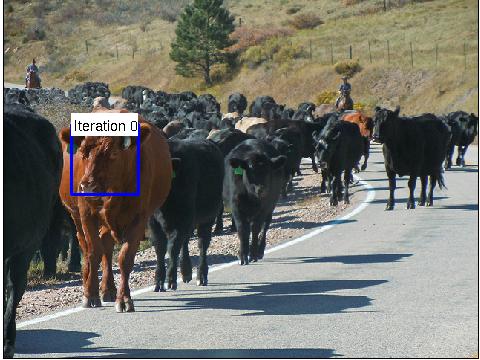}
\includegraphics[width=2.0cm]{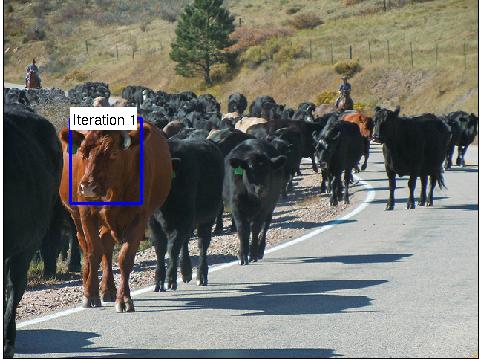}
\includegraphics[width=2.0cm]{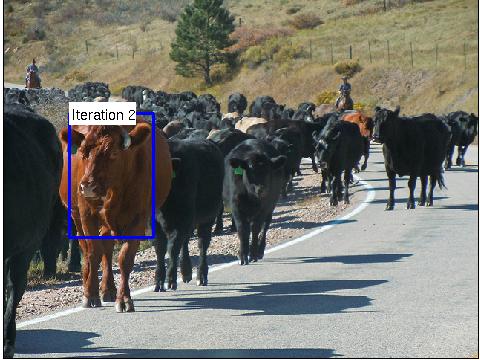}
\includegraphics[width=2.0cm]{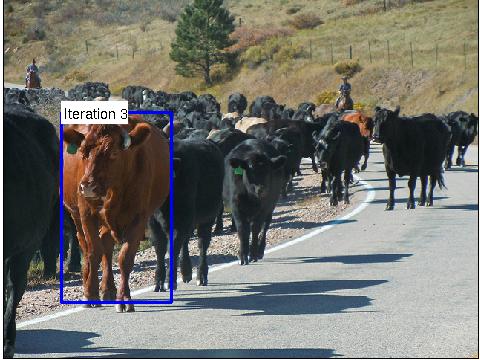}
\includegraphics[width=2.0cm]{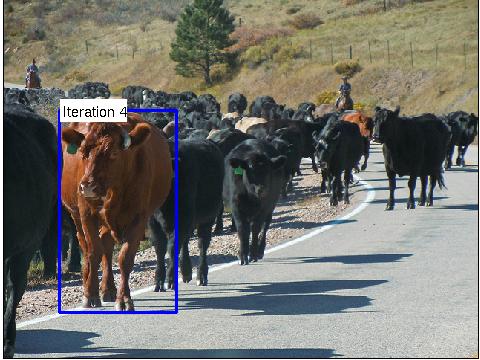}
\includegraphics[width=2.0cm]{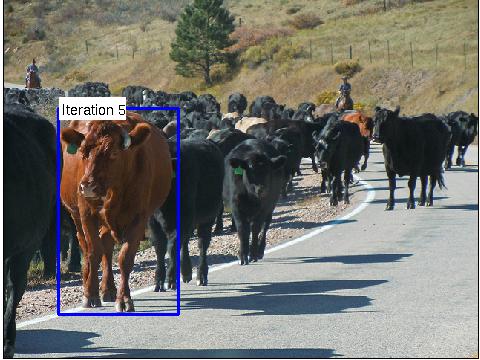}}\\
\vspace{-1pt}
\bmvaHangBox{
\includegraphics[width=2.0cm]{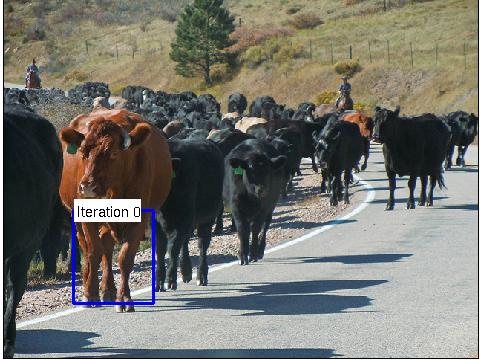}
\includegraphics[width=2.0cm]{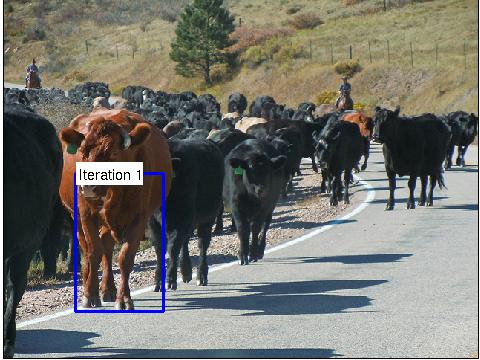}
\includegraphics[width=2.0cm]{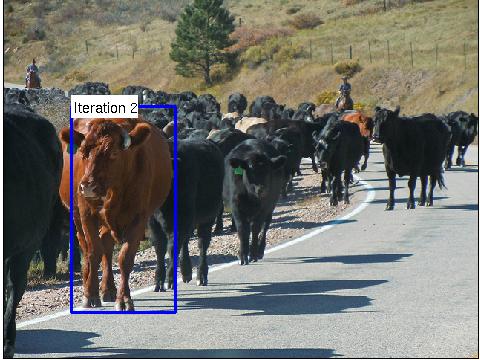}
\includegraphics[width=2.0cm]{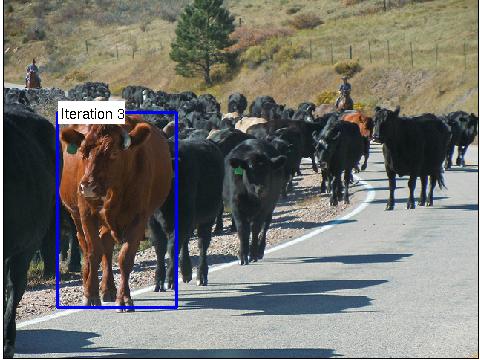}
\includegraphics[width=2.0cm]{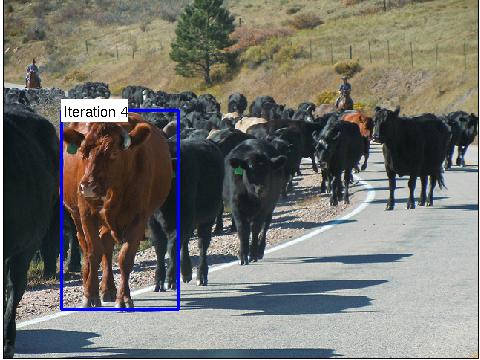}
\includegraphics[width=2.0cm]{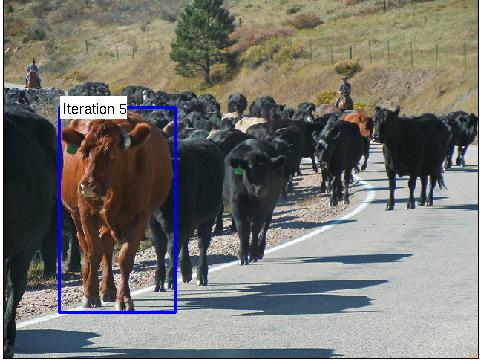}}\\
\vspace{-1pt}
\bmvaHangBox{
\includegraphics[width=2.0cm]{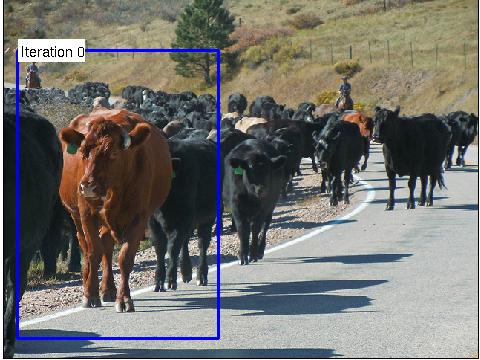}
\includegraphics[width=2.0cm]{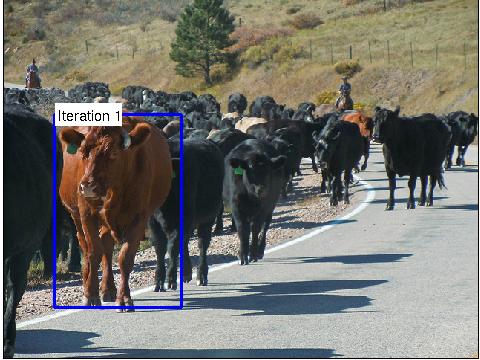}
\includegraphics[width=2.0cm]{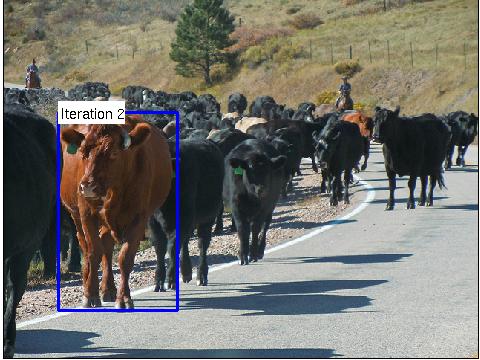}
\includegraphics[width=2.0cm]{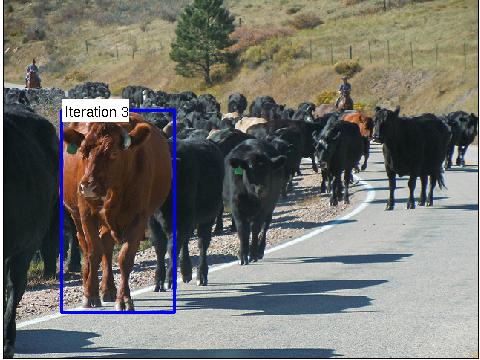}
\includegraphics[width=2.0cm]{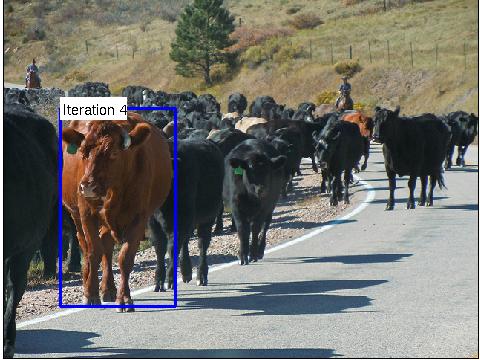}
\includegraphics[width=2.0cm]{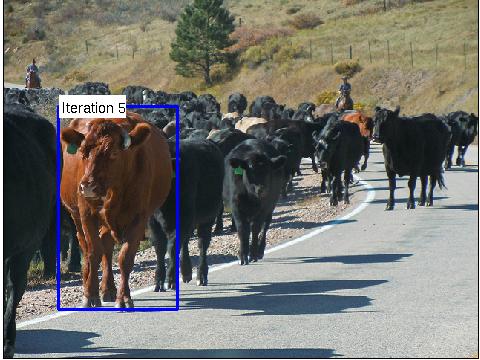}}\\
\vspace{-1pt}
\bmvaHangBox{
\includegraphics[width=2.0cm]{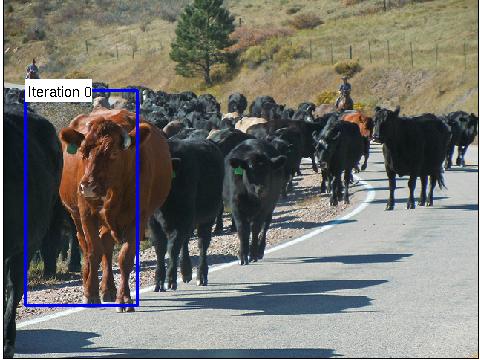}
\includegraphics[width=2.0cm]{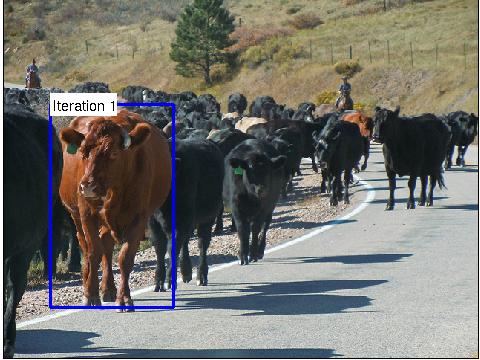}
\includegraphics[width=2.0cm]{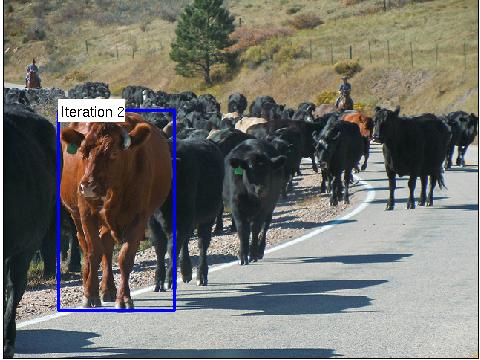}
\includegraphics[width=2.0cm]{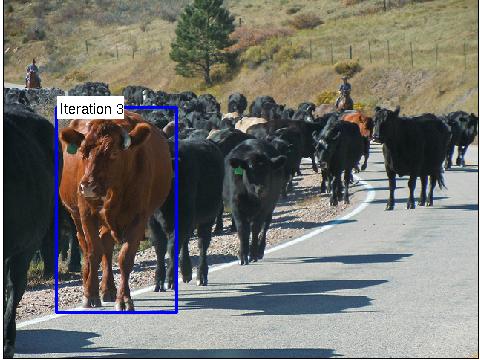}
\includegraphics[width=2.0cm]{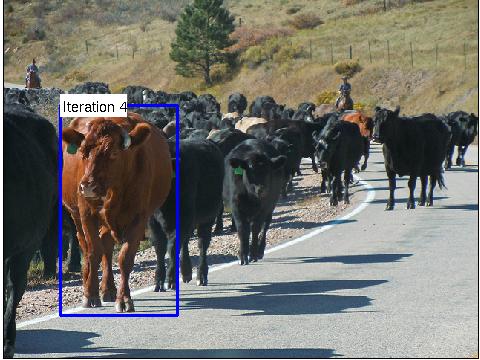}
\includegraphics[width=2.0cm]{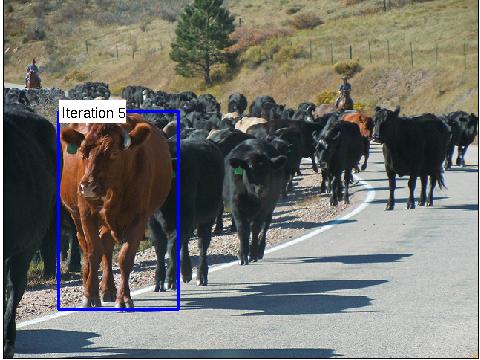}}\\
\end{center}
\vspace{-10pt}
\caption{\small{Illustration of the consecutive bounding box predictions made by our category agnostic location refinement module. In each row, from left to right we depict a seed box (iteration 0) and the bounding box predictions in each iteration. Despite the fact that the seed box might be quite far from the object (in terms of center location, scale and/or aspect ratio) the refinement module has no problem in converging to the bounding box  closest to the seed box object.
This capability is not affected even in the case that the seed box contains also other instances of the same category as in rows 3 and 4.}}
\label{fig:IterPreds}
\vspace{-15pt}
\end{figure}

The advantages of having an algorithm that sequentially generates new box locations given the predictions of the previous stage are two-fold: 
\begin{compactitem}

\item
\textbf{Attention mechanism:}
First, it behaves as an attention mechanism that, on each iteration, focuses more and more on the promising locations (in terms of box coordinates) of the image (see Figure~\ref{fig:Attentionmaps}). As a result of this,  boxes that tightly enclose the image objects are more likely to be generated and to be scored with high objectness confidence. 

\item
\textbf{Robustness to initial boxes:}
Furthermore, it allows to refine some initially imperfect box predictions or to localize objects that might be far (in terms of center location, scale and/or aspect ratio) from any seed box in the image. This is illustrated via a few characteristic examples in Figure~\ref{fig:IterPreds}. As shown in each of these examples, starting from a seed box, the iterative bounding box predictions   gradually converge to the closest (in terms of center location, scale and/or aspect ratio) object without actually being  affected from any nearby instances.
\end{compactitem}

\subsection{CNN-based box proposal model} \label{sec:model}

In this section we describe in more detail the object localization and objectness scoring modules of our box proposal model as well as the CNN architecture that implements the entire \emph{Attend Refine Repeat} algorithm that was presented above.
\vspace{-10pt}
\subsubsection{Object location refinement module} \label{sec:location_refinement}

In order for our active box proposals generation strategy to be effective, it is very important to have an accurate and robust category agnostic object location refinement module. 
Hence we follow the paradigm of the recently introduced LocNet model~\cite{gidaris2016locnet} that has demonstrated superior performance in the category specific object detection task over the typical bounding box regression paradigm~\cite{gidaris2015object,girshick2015fast,sermanet2013overfeat,shaoqing2015faster} by formulating the problem of bounding box prediction as a dense classification task. Here we use a properly adapted version of that model for the task at hand.

\begin{figure}
\begin{center}
\bmvaHangBox{
\includegraphics[width=4.0cm]{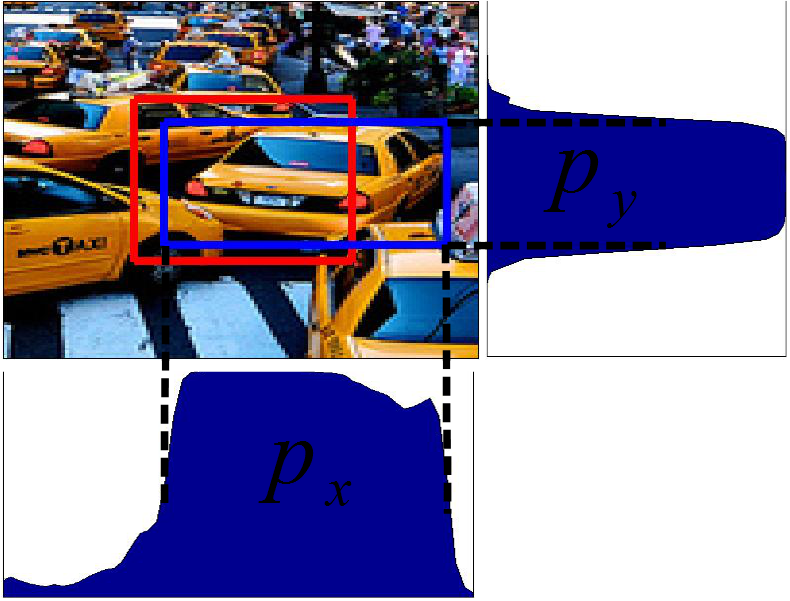}
\includegraphics[width=4.0cm]{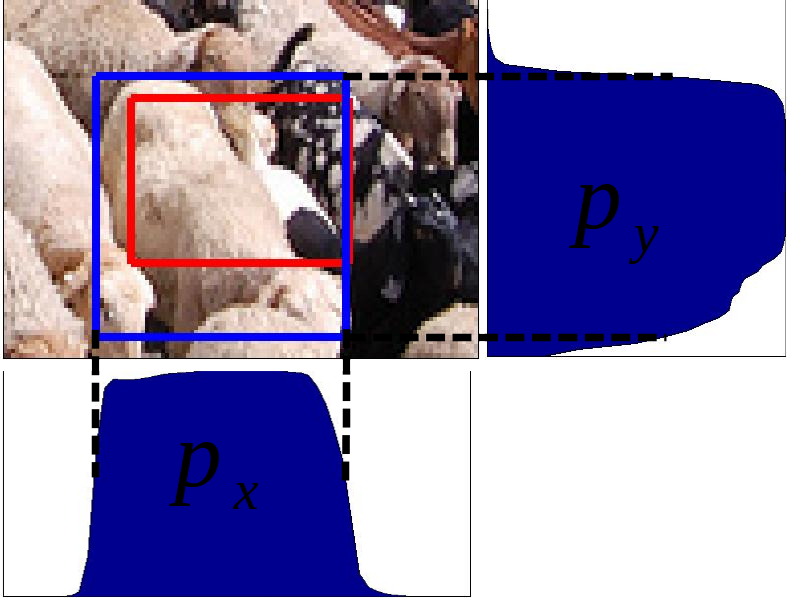}
\includegraphics[width=4.0cm]{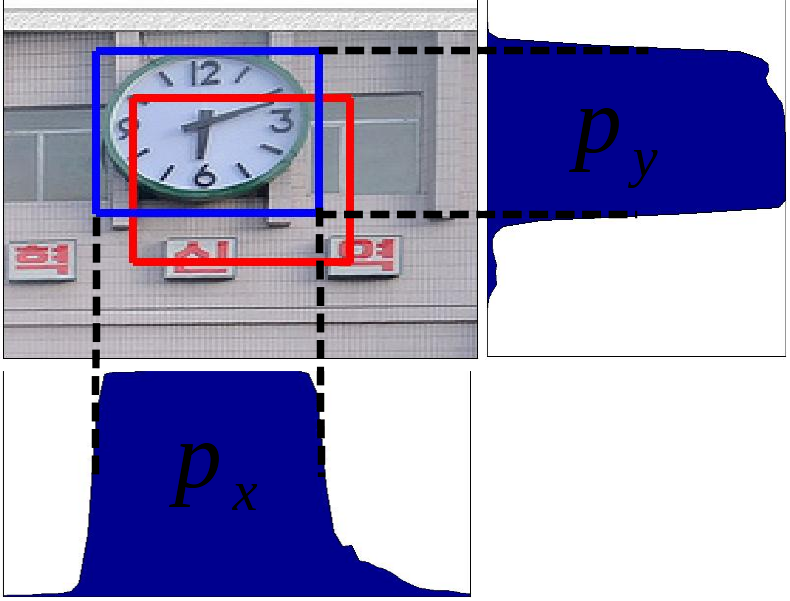}}\\
\end{center}
\vspace{-10pt}
\caption{\small{Illustration of the bounding box prediction process that is performed by our location refinement module.
In each case the red rectangle is the input box $B$, the blue rectangle is the predicted box and the depicted image crops are the search regions where the refinement module "looks" in order to localize the target object. 
On the bottom and on the right side of the image crop we visualize the $p_x$ and $p_y$ probability vectors respectively that our location refinement module yields in order to localize the target object.
Ideally, those probabilities should be equal to 1 for the elements (columns/rows) that overlap with the target box and 0 everywhere else. 
In the two left-most examples, the object location refinement module manages to correctly return the location of the target objects despite the fact that they are in crowded scenes with other objects of the same appearance and/or texture. In the right-most example, we qualitatively test the generalization capability of an location refinement module trained on PASCAL when applied to an object whose category ("clock") was not present in the training set.
}}
\label{fig:inout}
\vspace{-15pt}
\end{figure}

At a high level, given as input a bounding box $B$, the location refinement module first defines a search region $R = \gamma B$ (i.e., the region of $B$ enlarged by a factor $\gamma$) over which it is going to next search for a new refined bounding box. To achieve this, it  considers a discretization of  the search region  $R$ into $M$ columns as well as  $M$ rows, and  yields two probability vectors,  ${p}_{x} \!=\! \{p_{x,i}\}_{i=1}^{M}$ and ${p}_{y} = \{p_{y,i}\}_{i=1}^{M}$, for the $M$ columns and the $M$ rows respectively of $R$, where these probabilities  represent the likelihood of those elements (rows or columns) to be inside the target box $B^{*}$ (these are also called \emph{in-out} probabilities in the original LocNet model). Each time the target box $B^{*}$ is defined to be the bounding box of the object closest to the  input box $B$. 
Finally, given those \emph{in-out} probabilities, the object location $\tilde{B}$ inference is formulated as a simple maximum likelihood estimation problem that maximizes the likelihood of the \emph{in-out} elements of $\tilde{B}$. A visual illustration of the above process through a few examples is provided in Fig.~\ref{fig:inout} (for further details about the LocNet model we refer the interested reader to~\cite{gidaris2016locnet}).

We note that in contrast to the original LocNet model that is optimized to yield a different set of probability vectors of each category in the training set, here our category-agnostic version is designed to yield a single set of probability vectors that should accurately localize any object regardless of its category (see also section $\S$\ref{sec:architecture} that describes in detail the overall architecture of our proposed model).
It should be also mentioned that this is a more challenging task to learn since, in this case,  the model should be able to localize the target objects even if they are in crowded scenes with other objects of the same appearance and/or texture (see the two left-most examples of Figure~\ref{fig:inout}) without exploiting any category supervision during training that  would help it to better capture the appearance characteristics of each object category.
On top of that, our model should be able to localize objects of unseen categories. In the right-most example of Figure~\ref{fig:inout}, we provide an indicative result produced by our model that verifies this test case.
In this particular example, we apply a category-agnostic refinement module trained on PASCAL to an object whose category ("clock") was not present in the training set and yet our trained model had no problem of confidently predicting the correct location of the object. 
In section~\ref{sec:generalization} of the paper we also provide quantitative results about the generalization capabilities of the location refinement module.

\vspace{-10pt}
\subsubsection{Objectness scoring module} \label{sec:objectness}
The functionality of the objectness scoring module is that it gets as input a box $B$ and yields a single probability $p_{obj}$ of whether or not this box tightly encloses an object, regardless of what the category of that object might be. 

\subsubsection{AttractioNet architecture} \label{sec:architecture}
\begin{figure}[t]
\begin{center}
\bmvaHangBox{\includegraphics[width=0.75\textwidth]{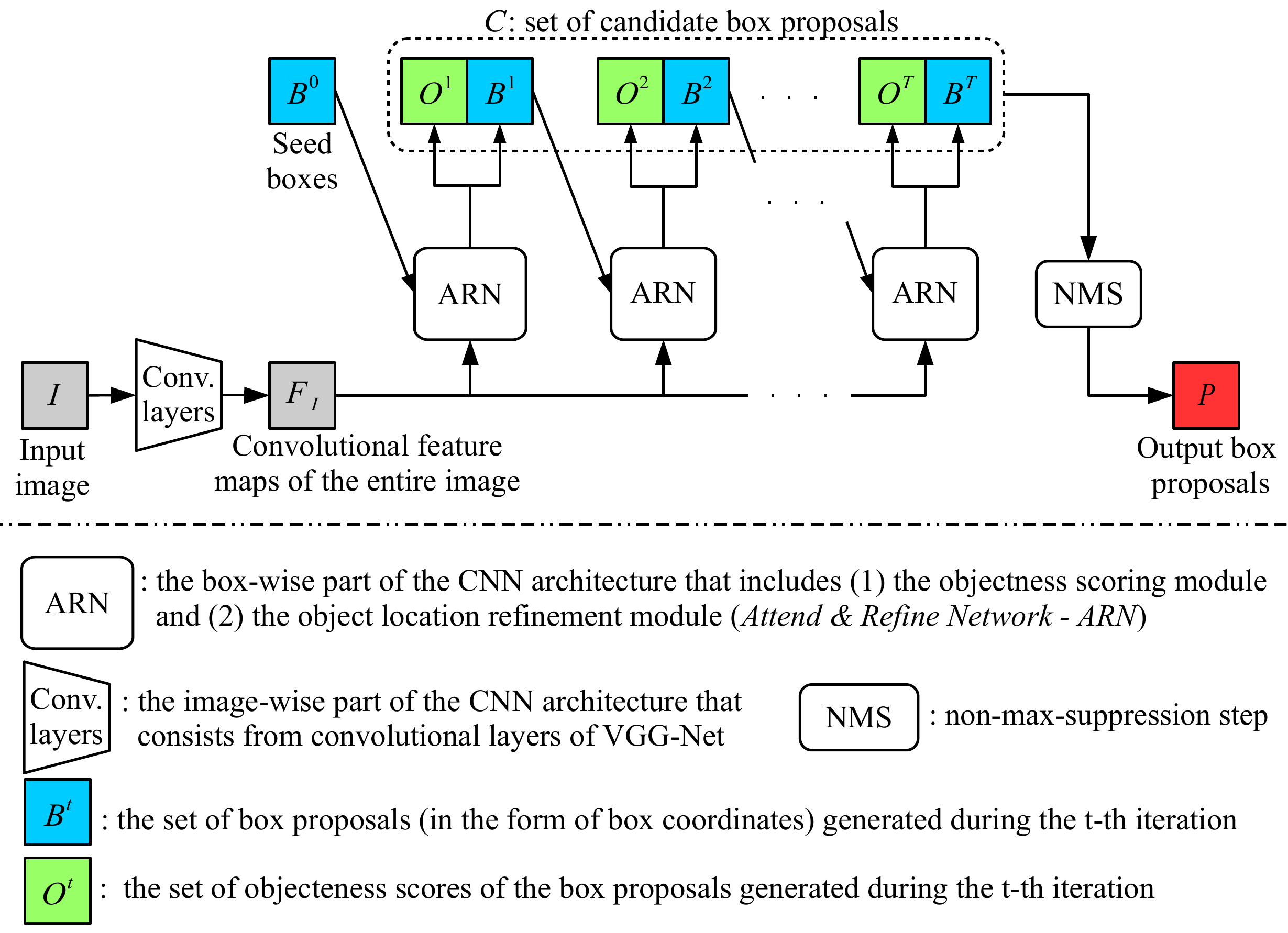}}
\end{center}
\vspace{-5pt}
\caption{\small{\textbf{\emph{AttractioNet work-flow}.} 
The \emph{Attend Refine Repeat} algorithm is implemented through a CNN model, called \emph{AttractioNet}, whose run-time work-flow (when un-rolled over time) is illustrated here.
On each iteration $t$ the box-wise part of the architecture (\emph{Attend \& Refine Network: ARN}) gets as input 
the image convolutional feature maps $F_I$ (extracted from the image-wise part of the CNN architecture)
as well as a set of box locations $\textbf{B}^{t-1}$ and yields the refined bounding box locations $\textbf{B}^{t}$ and their objectness scores $\textbf{O}^{t}$ using its \emph{category agnostic object location refinement} module and its \emph{category agnostic objectness scoring} module respectively. 
To avoid any confusion, note that our \emph{AttractioNet} model does not include any recurrent connections.
}}
\vspace{-15pt}
\label{fig:workflow}
\end{figure}
\begin{figure}[t]
\begin{center}
\bmvaHangBox{\includegraphics[width=\textwidth]{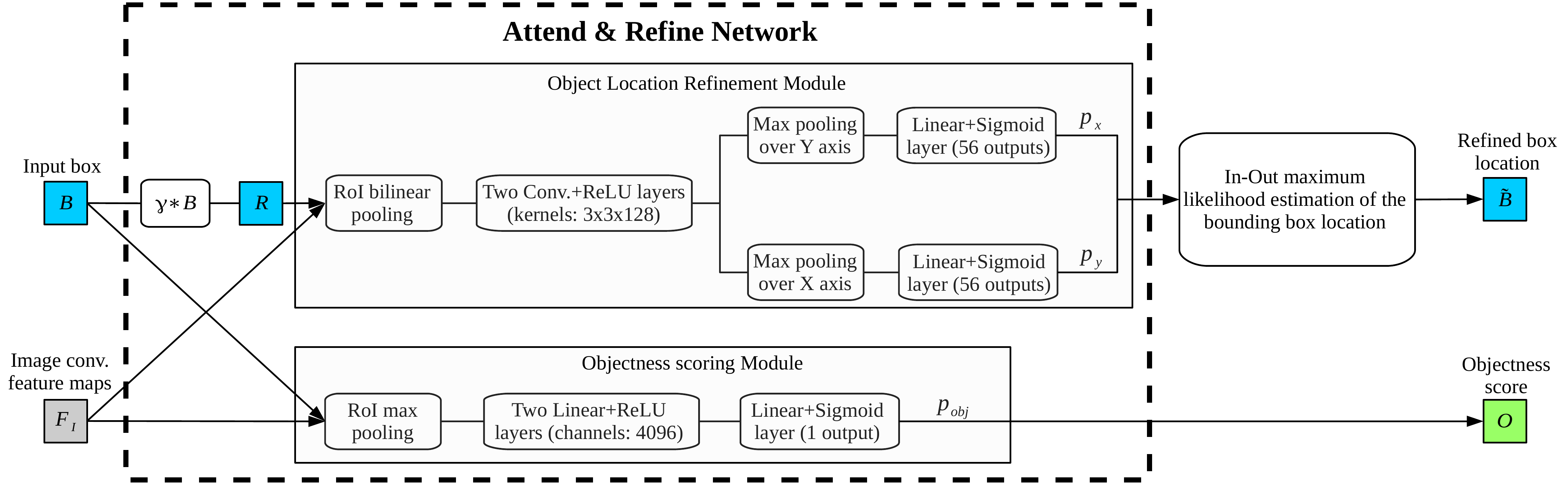}}
\end{center}
\vspace{-10pt}
\caption{\small{\textbf{\emph{Attend \& Refine Network} architecture.} The \emph{Attend \& Refine Network} is the box-wise part of the \emph{AttractioNet} architecture. In this figure we depict the work-flow for a single input box $B$.
Specifically, given an input box $B$ and the image convolutional feature maps $F_I$, the \emph{Attend \& Refine Network} yields \textbf{(1)} the in-out location probability vectors, $p_x$ and $p_y$, (using its object location refinement sub-network) and \textbf{(2)} the objectness scalar probability $p_{obj}$ (using its objectness scoring sub-network). 
Given the \emph{in-out} probabilities, $p_x$ and $p_y$, the object location inference is formulated as a simple maximum likelihood estimation problem that results on the refined bounding box coordinates $\tilde{B}$.}}
\label{fig:ARN}
\vspace{-10pt}
\end{figure}

We call the overall network architecture that implements the \emph{Attend Refine Repeat} algorithm with its \emph{In-Out} object location refinement module and its objectness scoring module, \emph{AttractioNet}\footnote{\emph{AttractioNet : (Att)end (R)efine Repeat: (Act)ive Box Proposal Generation via (I)n-(O)ut Localization (Net)work}}.
Given an image $I$, our \emph{AttractioNet} model will be required to process multiple image boxes of various sizes, by two different modules and repeat those processing steps for several iterations of the \emph{Attend Refine Repeat} algorithm. 
So, in order to have an efficient implementation we follow the SPP-Net~\cite{he2015spatial} and Fast-RCNN~\cite{girshick2015fast} paradigm and share the operations of the first convolutional layers between all the boxes, as well as across the two modules and all the \emph{Attend Refine Repeat} algorithm repetitions (see Figure~\ref{fig:workflow}). 
Specifically, our \emph{AttractioNet} model first forwards the image $I$ through a first sequence of convolutional layers (conv. layers of VGG16-Net~\cite{simonyan2014very}) in order to extract convolutional feature maps $F_I$ from the entire image. 
Then, on each iteration $t$ the box-wise part of the architecture, which we call \emph{Attend \& Refine Network},
gets as input the image convolutional feature maps $F_I$ 
and a set of box locations $\textbf{B}^{t-1}$ and yields the refined bounding box locations $\textbf{B}^{t}$ and their objectness scores $\textbf{O}^{t}$ using its object location refinement module sub-network and its objectness scoring module sub-network respectively.
In Figure~\ref{fig:ARN} we provide the work-flow of the \emph{Attend \& Refine Network} when processing a single input box $B$.
The architecture of its two sub-networks is described in more detail in the rest of this section:\\

\textbf{Object location refinement module sub-network.} 
This module gets as input the feature map $F_I$ and the search region $R$ and yields the probability vectors $p_x$ and $p_y$ of that search region with a network architecture similar to that of LocNet. 
Key elements of this architecture is that it branches into two heads, the X and Y, each responsible for yielding the $p_x$ or the $p_y$ outputs.  
Differently from the original LocNet architecture, the convolutional layers of this sub-network output 128 feature channels instead of 512, which speeds up the processing by a factor of 4 without affecting the category-agnostic localization accuracy.
Also, in order to yield a fixed size feature for the $R$ region, instead of region adaptive max-pooling this sub-network uses region bilinear pooling~\cite{dai2015instance,densecap} that in our initial experiments gave slightly better results.
Finally, 
our version is designed to yield two probability vectors of size $M$\footnote{Here we use  $M=56$.}, instead of $C \times 2$ vectors of size $M$ 
(where $C$ is the number of categories), given that in our case we aim for category-agnostic object location refinement.\\

\textbf{Objectness scoring module sub-network.} 
Given the image feature maps $F_I$ and the window $B$ it first performs region adaptive max pooling of the features inside $B$ that yields a fixed size feature ($7 \times 7 \times 512$).
Then it forwards this feature through two linear+ReLU hidden layers of $4096$ channels each (fc\_6 and fc\_7 layers of VGG16) and a final linear+sigmoid layer with one output that corresponds to the probability $p_{obj}$ of the box $B$ tightly enclosing an object. During training the hidden layers are followed by Dropout units with dropout probability $p=0.5$.\\
\vspace{-15pt}
\subsection{Training procedure} \label{sec:training}
\textbf{Training loss:} During training the following multi-task loss is optimized:
\begin{equation} \label{eq:total_loss_log}
\underbrace{\frac{1}{N^L} \sum_{k=1}^{N^L}{L_{loc}(\theta| B_k, T_k, I_k)}}_\text{localization task loss} + \underbrace{\frac{1}{N^O} \sum_{k=1}^{N^O}{L_{obj}(\theta| B_k, y_k, I_k)}}_\text{objectness scoring task loss} \textbf{,}
\end{equation}
where $\theta$ are the learnable network parameters, $\{B_k, T_k, I_k\}_{k=1}^{N^L}$ are $N^L$ training triplets for learning the localization task and $\{B_k, y_k, I_k\}_{k=1}^{N^O}$ are $N^O$ training triplets for learning the objectness scoring task. 
Each training triple $\{B, T, I\}$ of the localization task includes the image $I$, the box $B$ and the target localization probability vectors $T=\{T_x,T_y\}$. 
If $(B^{*}_l, B^{*}_t)$ and $(B^{*}_r, B^{*}_b)$ are the top-left and bottom-right coordinates of the target box $B^{*}$ then the target probability vectors ${T}_{x} \!=\! \{T_{x,i}\}_{i=1}^{M}$ and ${T}_{y} \!=\! \{T_{y,i}\}_{i=1}^{M}$ are defined as:
\begin{align}  \label{eq:targets_loc}
  T_{x,i} =\left\{
  \begin{array}{@{}ll@{}}
    1, & \text{if}\ B^{*}_l \leq i \leq B^{*}_r \\
    0, & \text{otherwise}
  \end{array}\right.\text{and  }
  T_{y,i}=\left\{
  \begin{array}{@{}ll@{}}
    1, & \text{if}\ B^{*}_t \leq i \leq B^{*}_b \\
    0, & \text{otherwise}
  \end{array}\right.\text{, } \forall i \in \{1,\ldots,M\}
\end{align} 
The loss $L_{loc}(\theta| B, T, I)$ of this triplet is the sum of binary logistic regression losses:
\begin{equation} \label{eq:inout_loss_log}
\frac{1}{2M} \sum_{a \in\\ \{x,y\}}\sum_{i=1}^M T_{a,i}\log(p_{a,i})+(1-T_{a,i})\log(1 - p_{a,i})\,,
\end{equation}
where $p_{a}$ are the output probability vectors of the localization module for the image $I$, the box $B$ and the network parameters $\theta$.
The training triplet $\{B, y, I\}$ for the objectness scoring task includes the image $I$, the box $B$ and the target value $y \in \{0,1\}$ of whether the box $B$ contains an object (positive triplet with $y = 1$) or not (negative triplet with $y = 0$). 
The loss $L_{obj}(\theta| B, y, I)$ of this triplet is the binary logistic regression loss $y\log(p_{obj})+(1-y)\log(1 - p_{obj})$,
where $p_{obj}$ is the objectness probability for the image $I$, the box $B$ and the network parameters $\theta$.\\

\textbf{Creating training triplets:} 
In order to create the localization and objectness training triplets of one image we first artificially create a pool of boxes that our algorithm is likely to see during  test time. 
Hence we start by generating seed boxes (as the test time algorithm) and for each of them we predict the bounding boxes of the ground truth objects that are closest to them using an ideal object location refinement module.
This step is repeated one more time using the previous ideal predictions as input. 
Because of the finite search area of the search region $R$ the predicted boxes will not necessarily coincide with the ground truth bounding boxes. 
Furthermore, to account for prediction errors during test time, we repeat the above process by jittering this time the output probability vectors of the ideal location refinement module with $20\%$ noise.
Finally, we merge all the generated boxes (starting from the seed ones) to a single pool. 
Given this pool, the positive training boxes in the objectness localization task are those that their $IoU$ with any ground truth object is at least $0.5$ and the negative training boxes are those that their maximum $IoU$ with any ground truth object is less than $0.4$. For the localization task we use as training boxes those that their $IoU$ with any ground truth object is at least $0.5$.\\

\textbf{Optimization:} To minimize the objective we use stochastic gradient descent (SGD) optimization with an image-centric strategy for sampling training triplets. 
Specifically, in each mini-batch we first sample 4 images and then for each image we sample 64 training triplets for the objectness scoring task ($50\%$ are positive and $50\%$ are negative) and 32 training triplets for the localization task. 
The momentum is set to $0.9$ and the learning schedule includes training for $320k$ iterations with a learning rate of $l_r=0.001$ and then for another $260k$ iterations with $l_r=0.0001$. 
The training time is around 7 days (although we observed that we could have stopped training on the 5th day with insignificant loss in performance).\\

\textbf{Scale and aspect ratio jittering:}
During test time our model is fed with a single image scaled such that its shortest dimension to be $1000$ pixels or its longest dimension to not exceed the $1400$ pixels.
However, during training each image is randomly resized such that its shortest dimension to be one of the following number of pixels $\{300\!:\!50\!:\!1000\}$ (using Matlab notation) taking care, however,  the longest dimension to not exceed  $1000$ pixels. 
Also, with probability $0.5$ we jitter the aspect ratio of the image by altering the image dimensions from $W \times H$ to $(\alpha W) \times H$ or $W \times (\alpha H)$ where the value of $\alpha$ is uniformly sampled from $2^{.2:.2:1.0}$ (Matlab notation). We observed that this type of data augmentation gives a slight improvement on the results.

\section{Experimental results} \label{sec:results}

In this section we perform an exhaustive evaluation of our box proposal generation approach, which we call \emph{AttractioNet}, under various test scenarios. 
Specifically, we first evaluate our approach with respect to its  object localization performance by comparing it with other competing methods and we also provide an ablation study of its main novel components in~\S\ref{sec:prop_gen_eval}.
Then, we study its ability to generalize to unseen categories in~\S\ref{sec:generalization}, 
we evaluate it in the context of the object detection task in~\S\ref{sec:detection} and 
finally, we provide qualitative results in~\S\ref{sec:qualitative}.\\

\textbf{Training set:} 
In order to train our \emph{AttractioNet} model we use the training set of MS COCO~\cite{lin2014microsoft} detection benchmark dataset that includes $80k$ images and it is labelled with $80$ different object categories.
Note that the MSCOCO dataset is an ideal candidate for training our box proposal model since: 
\textbf{(1)} it is labelled with a descent number of different object categories and
\textbf{(2)} it includes images captured from complex real-life scenes containing common objects in their natural context. 
The aforementioned training set properties are desirable for achieving good performance on "difficult" test images (a.k.a. images in the wild) and generalizing to "unseen" during training object categories.\\

\textbf{Implementation details:}
In the active box proposal algorithm we use $10k$ seed boxes generated with a similar to Cracking Bing~\cite{zhao2014cracking} technique\footnote{We use seed boxes of 3 aspect ratios, $1:2$, $2:1$ and $1:1$, and 9 different sizes of the smallest seed box dimension $\{16, 32, 50, 72, 96, 128, 192, 256, 384\}$.}. To reduce the computational cost of our algorithm, after the first repetition we only keep the top $2k$ scored boxes and we continue with this number of candidate box proposals for four more extra iterations.  
In the non-maximum-suppression~\cite{felzenszwalb2010object} (NMS) step the optimal IoU threshold (in terms of the achieved AR) depends on the desired number of box-proposals.
For example, for 10, 100, 1000 and 2000 proposals the optimal IoU thresholds are $0.55$, $0.75$, $0.90$ and $0.95$ respectively (note that the aforementioned IoU thresholds were cross validated on a set different from the one used for evaluation). 
For practical purposes and in order to have a unified NMS process, 
we first apply NMS with the IoU threshold equal to $0.95$ and get the top 2000 box proposals,
and then follow a multi-threshold NMS strategy that re-orders this set of 2000 boxes such that 
for any given number $K$, the top $K$ box proposals in the set better cover (in terms of achieved AR) the objects in the image (see appendix~\ref{sec:nms}).\\

\vspace{-10pt}
\subsection{Object box proposal generation evaluation} \label{sec:prop_gen_eval}

Here we evaluate our \emph{AtractioNet} method in the end task of box proposal generation.
For that purpose, we test it on the first $5k$ images of the COCO validation set and the PASCAL~\cite{everingham2010pascal} VOC2007 test set (that also includes around $5k$ images).\\

\textbf{Evaluation Metrics:}
As evaluation metric we use the average recall (AR) which, for a fixed number of box proposals, averages the recall of the localized ground truth objects for several Intersection over Union (IoU) thresholds in the range .5:.05:.95 (Matlab notation).
The average recall metric has been proposed from Hosang \etal~\cite{hosang2015makes,Hosang2014Bmvc} 
where in their work they demonstrated that it correlates well with the average precision performance of box proposal based object detection systems.
In our case, in order to evaluate our method we report the AR results for 10, 100 and 1000 box proposals using the notation \emph{AR@10}, \emph{AR@100} and \emph{AR@1000} respectively. 
Also, in the case of 100 box proposals we also report the AR of the small ($\alpha < 32^2$), medium ($32^2 \leq \alpha \leq 96^2$) and large ($\alpha > 96^2$) sized objects using the notation \emph{AR@100-Small}, \emph{AR@100-Medium} and \emph{AR@100-Large} respectively, where $\alpha$ is the area of the object. For extracting those measurements we use the COCO API (\url{https://github.com/pdollar/coco}).\\

\subsubsection{Average recall evaluation} \label{sec:ar_results}
\begin{figure}[t!]
\begin{center}
\begin{tabular}{cc}
\bmvaHangBox{\includegraphics[width=5cm]{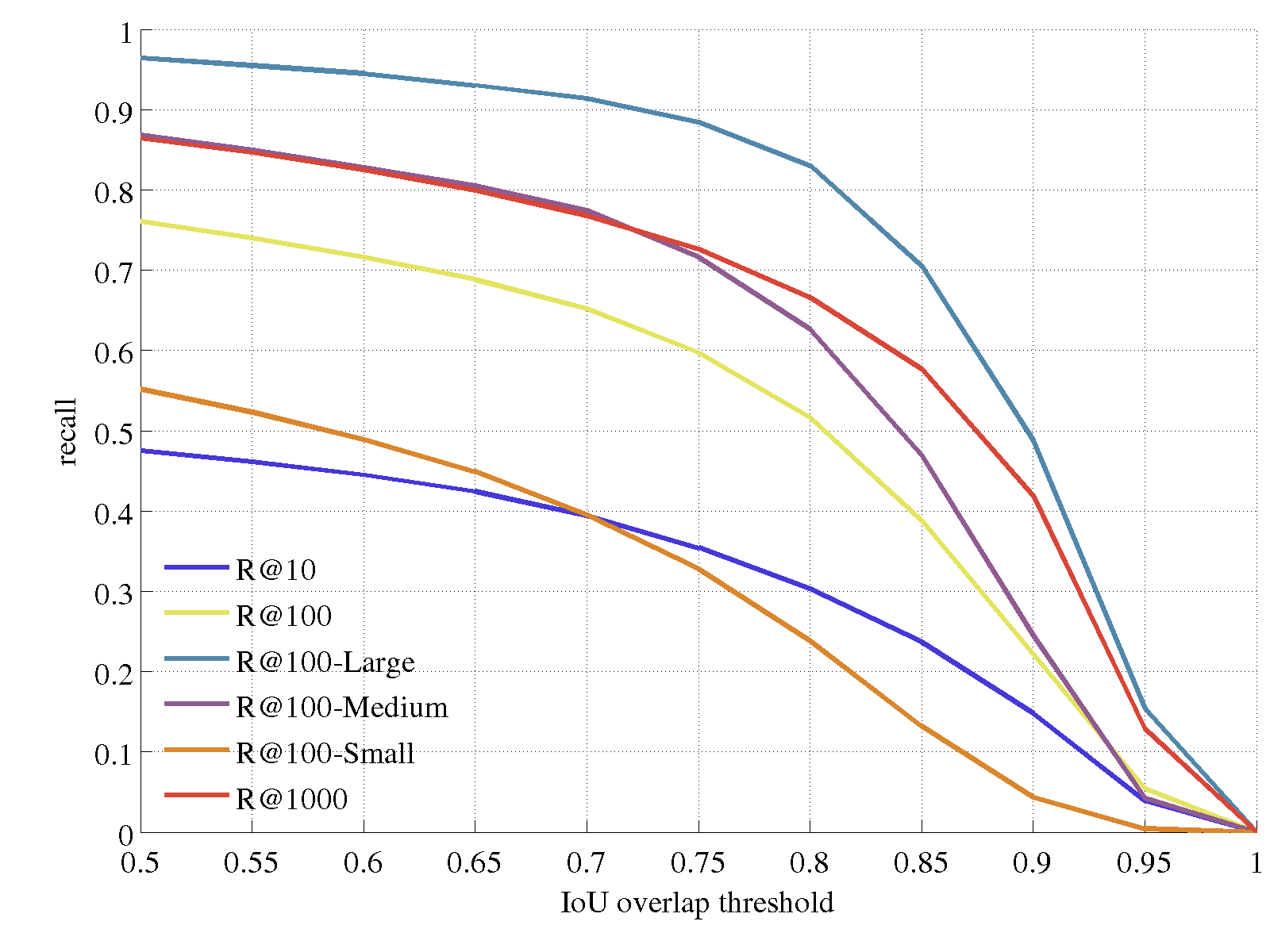}}&
\bmvaHangBox{\includegraphics[width=5cm]{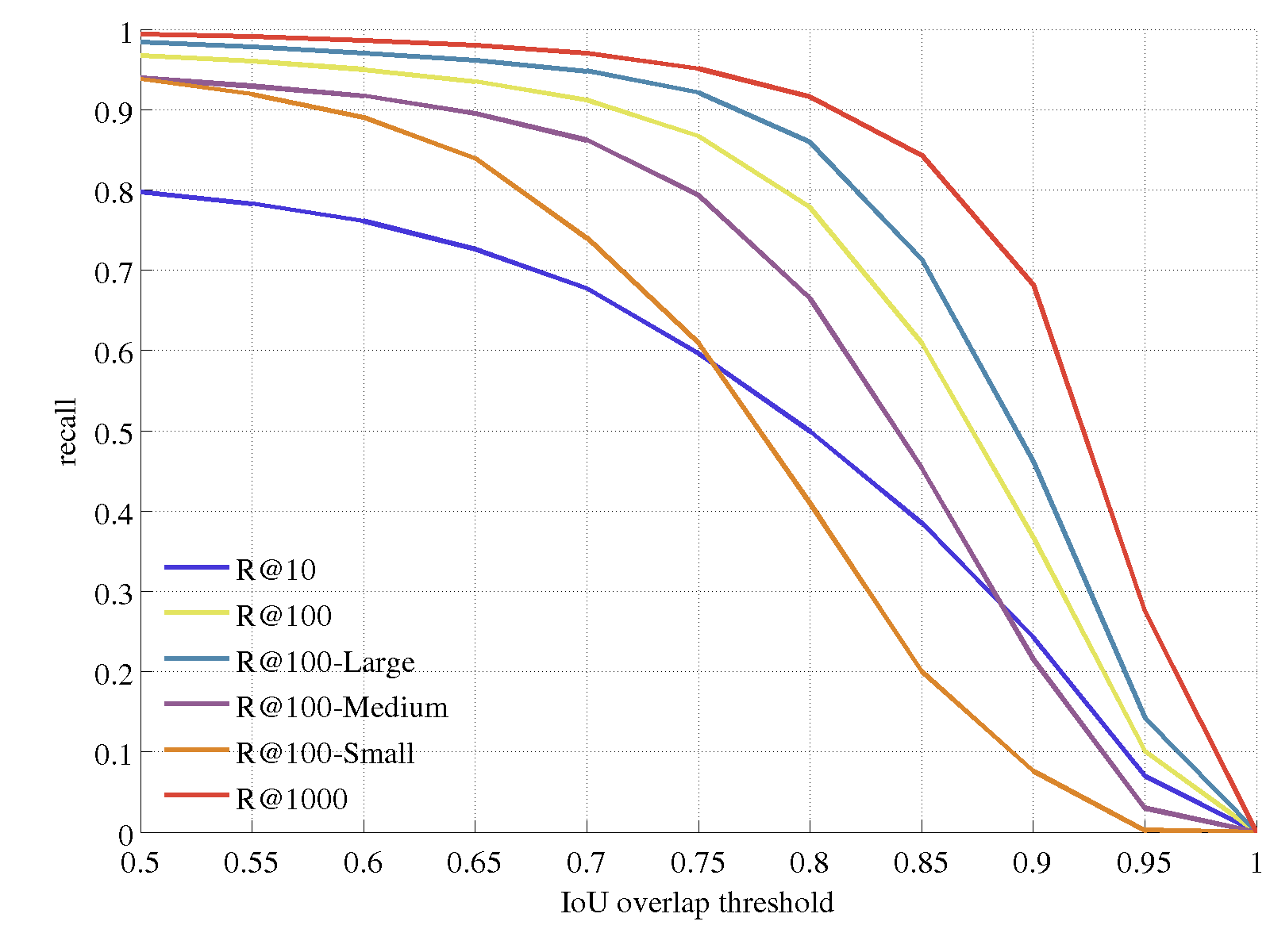}}\\
\end{tabular}
\end{center}
\vspace{-8pt}
\caption{\small{Recall versus IoU overlap plots of our \emph{AttractioNet} approach under different test cases: 10 proposals (\emph{R@10}), 100 proposals (\emph{R@100}), 1000 proposals (\emph{R@1000}), 100 proposals and small sized objects (\emph{R@100-Small}), 100 proposals and medium sized objects (\emph{R@100-Medium}) and  100 proposals and large sized objects (\emph{R@100-Large}). \textbf{(Left)}  Results in the first $5k$ images of COCO validation set. \textbf{(Right)} Results in the PASCAL VOC2007 test set.}}
\label{fig:recall_vs_iou}
\vspace{-10pt}
\end{figure}

In Table~\ref{tab:AR_COCO} we report the average recall (AR) metrics of our method as well as of other competing methods in the COCO validation set.
We observe that the average recall performance achieved by our method exceeds all the previous work in all the AR metrics  
by a significant margin (around 10 absolute points in the percentage scale).
Similar gains are also observed in Table~\ref{tab:AR_PASCAL} where we report the average recall results of our methods in the PASCAL VOC2007 test set.
Furthermore, in Figure~\ref{fig:recall_vs_iou} we provide for our method the recall as a function of the IoU overlap of the localized ground truth objects.
We see that the recall decreases relatively slowly as we increase the IoU from 0.5 to 0.75 while for IoU above 0.85 the decrease is faster. \\
\begin{table}[t!]
\centering
\resizebox{\textwidth}{!}{
{\setlength{\extrarowheight}{2pt}\scriptsize
\begin{tabular}{l | c c c c c c  }
\hline
Method                  & AR@10 & AR@100 & AR@1000 & AR@100-Small & AR@100-Medium & AR@100-Large\\
\hline
EdgeBoxes~\cite{zitnick2014edge}                & 0.074 & 0.178 & 0.338 & 0.015 & 0.134 & 0.502\\
Geodesic~\cite{krahenbuhl2014geodesic}      & 0.040 & 0.180 & 0.359 & - & - & -\\
Selective Search~\cite{van2011segmentation} & 0.052 & 0.163 & 0.357 & 0.012 & 0.0132 & 0.466\\
MCG~\cite{APBMM2014}                                    & 0.101 & 0.246 & 0.398 & 0.008 & 0.119 & 0.530\\
DeepMask~\cite{pinheiro2015learning}        & 0.153 & 0.313 & 0.446 & - & - & -\\
DeepMaskZoom~\cite{pinheiro2015learning}    & 0.150 & 0.326 & 0.482 & - & - & -\\
Co-Obj~\cite{hayder2016learning}                        & 0.189 & 0.366 & 0.492 & 0.107 & 0.449 & 0.686\\
SharpMask~\cite{PinheiroLCD16}                  & 0.192 & 0.362 & 0.483 & 0.060 & 0.510 & 0.665\\
SharpMaskZoom~\cite{PinheiroLCD16}              & 0.192 & 0.390 & 0.532 & 0.149 & 0.507 & 0.630\\
SharpMaskZoom$^2$~\cite{PinheiroLCD16}          & 0.178 & 0.391 & 0.555 & 0.221 & 0.454 & 0.588\\
\hline
AttractioNet (Ours)  & \textbf{0.328} & \textbf{0.533} & \textbf{0.662} & \textbf{0.315} & \textbf{0.622} & \textbf{0.777}\\
\hline
\end{tabular}}}
\vspace{-8pt}
\caption{\small{Average Recall results on the first $5k$ images of COCO validation set.}}
\label{tab:AR_COCO}
\vspace{-5pt}
\end{table}

\begin{table}[t!]
\centering
\resizebox{\textwidth}{!}{
{\setlength{\extrarowheight}{2pt}\scriptsize
\begin{tabular}{l | c c c c c c  }
\hline
Method                  & AR@10 & AR@100 & AR@1000 & AR@100-Small & AR@100-Medium & AR@100-Large\\
\hline
EdgeBoxes~\cite{zitnick2014edge}       & 0.203 & 0.407 & 0.601 & 0.035 & 0.159 & 0.559\\
Geodesic~\cite{krahenbuhl2014geodesic}       & 0.121 & 0.364 & 0.596 & - & - & -\\
Selective Search~\cite{van2011segmentation} & 0.085 & 0.347 & 0.618 & 0.017 & 0.134 & 0.364\\
MCG~\cite{APBMM2014}             & 0.232 & 0.462 & 0.634 & 0.073 & 0.228 & 0.618\\
DeepMask~\cite{pinheiro2015learning}        & 0.337 & 0.561 & 0.690 & - & - & -\\
Best of Co-Obj~\cite{hayder2016learning} & 0.430 & 0.602 & 0.745 & 0.453 & 0.517 & 0.654\\
\hline
AttractioNet (Ours) & \textbf{0.554} & \textbf{0.744} & \textbf{0.859} & \textbf{0.562} & \textbf{0.670} & \textbf{0.794}\\
\hline
\end{tabular}}}
\vspace{-8pt}
\caption{\small{Average Recall results on the PASCAL VOC2007 test set.}}
\label{tab:AR_PASCAL}
\vspace{-5pt}
\end{table}

\begin{figure}[t!]
\begin{center}
\resizebox{\textwidth}{!}{
{\setlength{\extrarowheight}{2pt}\scriptsize
\begin{tabular}{ccc}
\bmvaHangBox{\includegraphics[width=4.0cm]{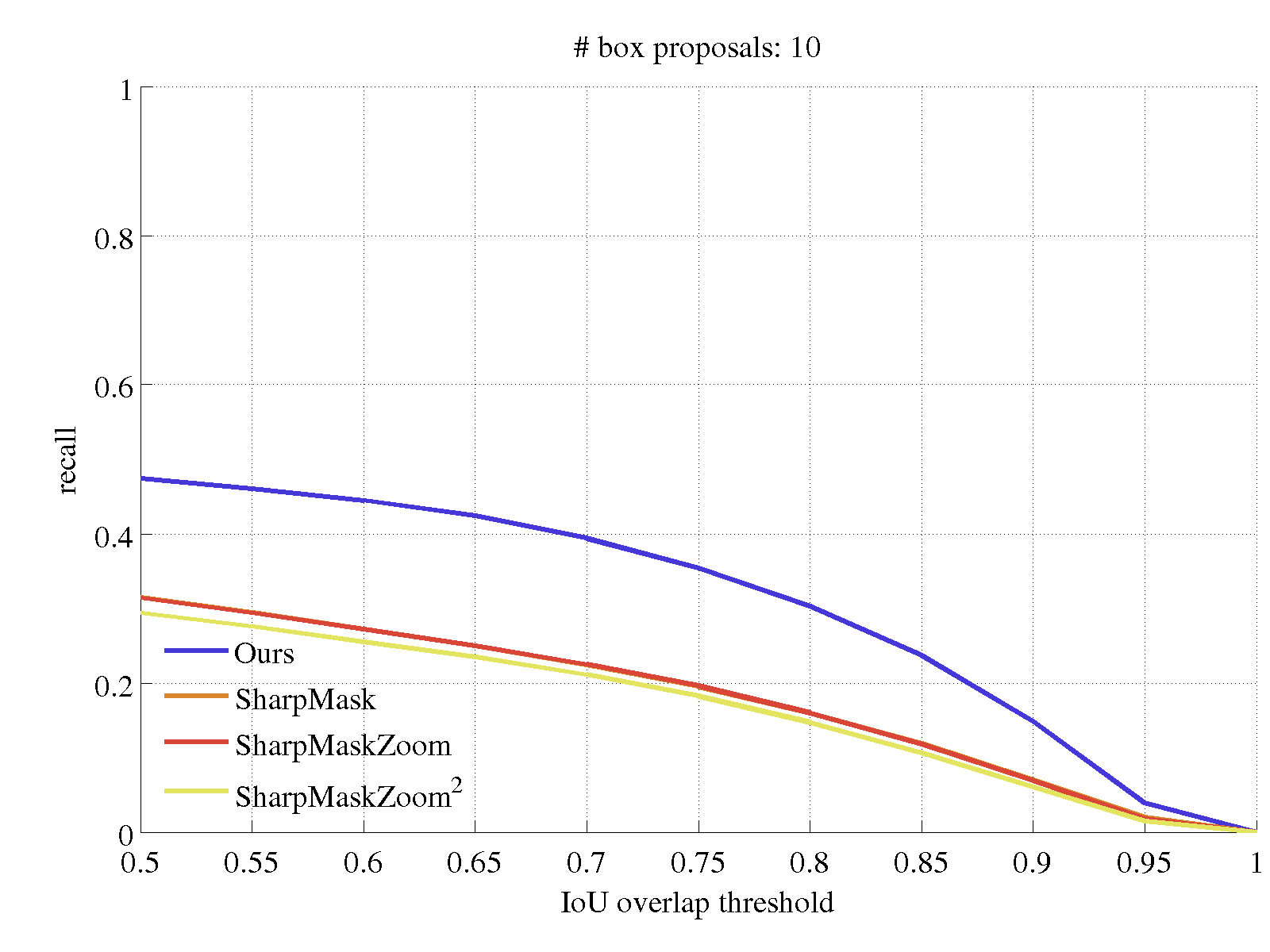}}&
\bmvaHangBox{\includegraphics[width=4.0cm]{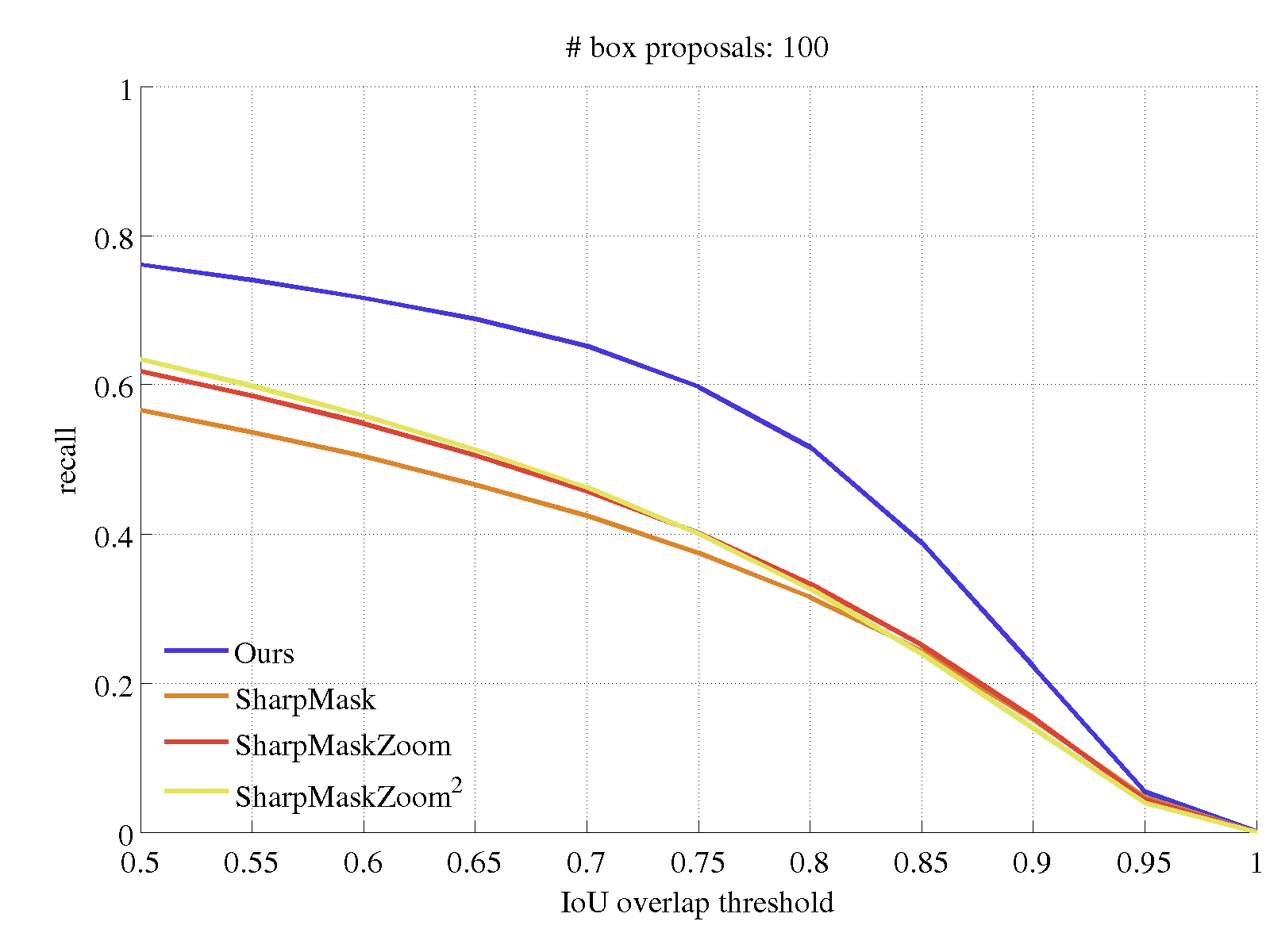}}&
\bmvaHangBox{\includegraphics[width=4.0cm]{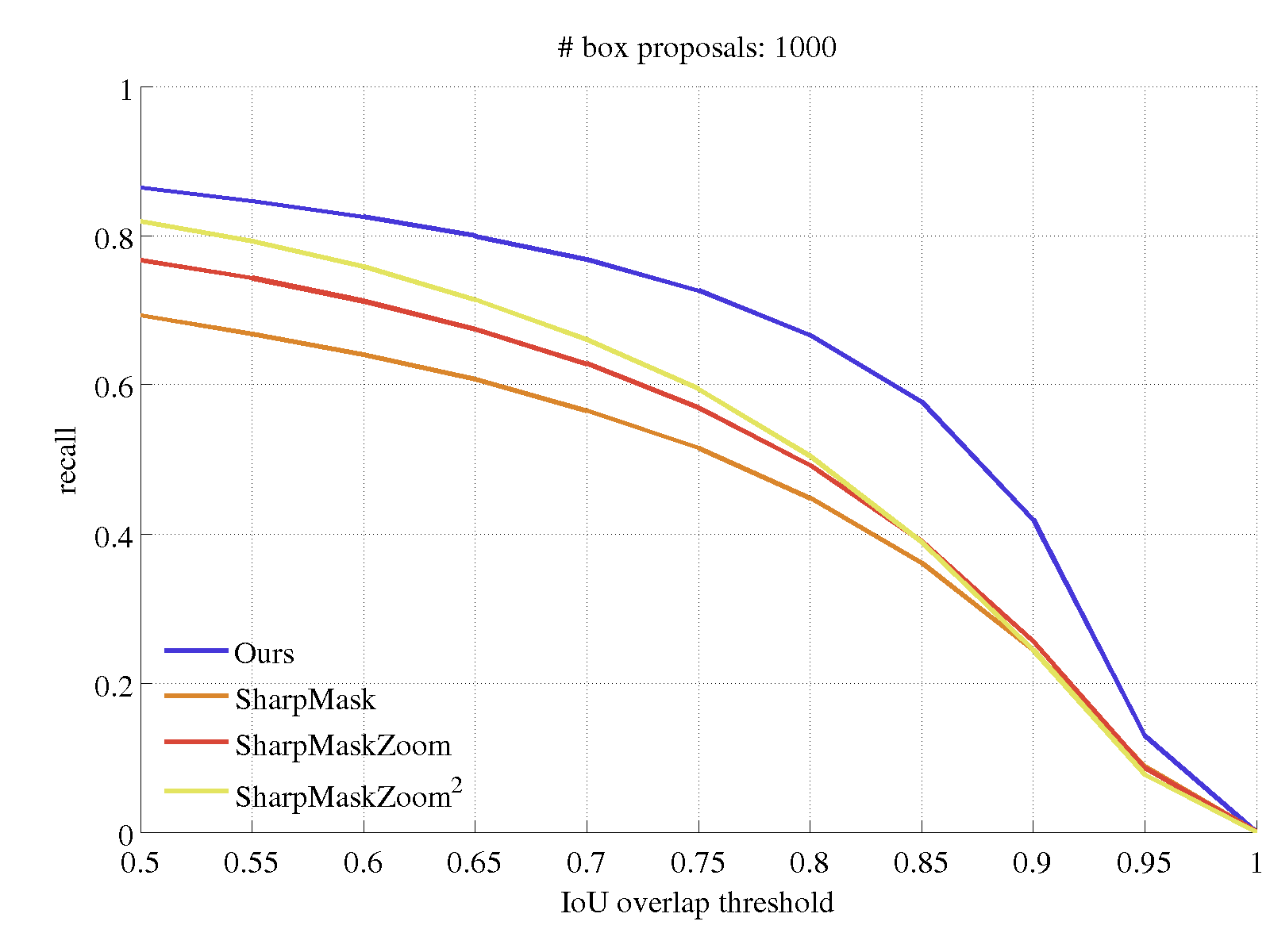}}\\
(a)&(b)&(c)\\
\bmvaHangBox{\includegraphics[width=4.0cm]{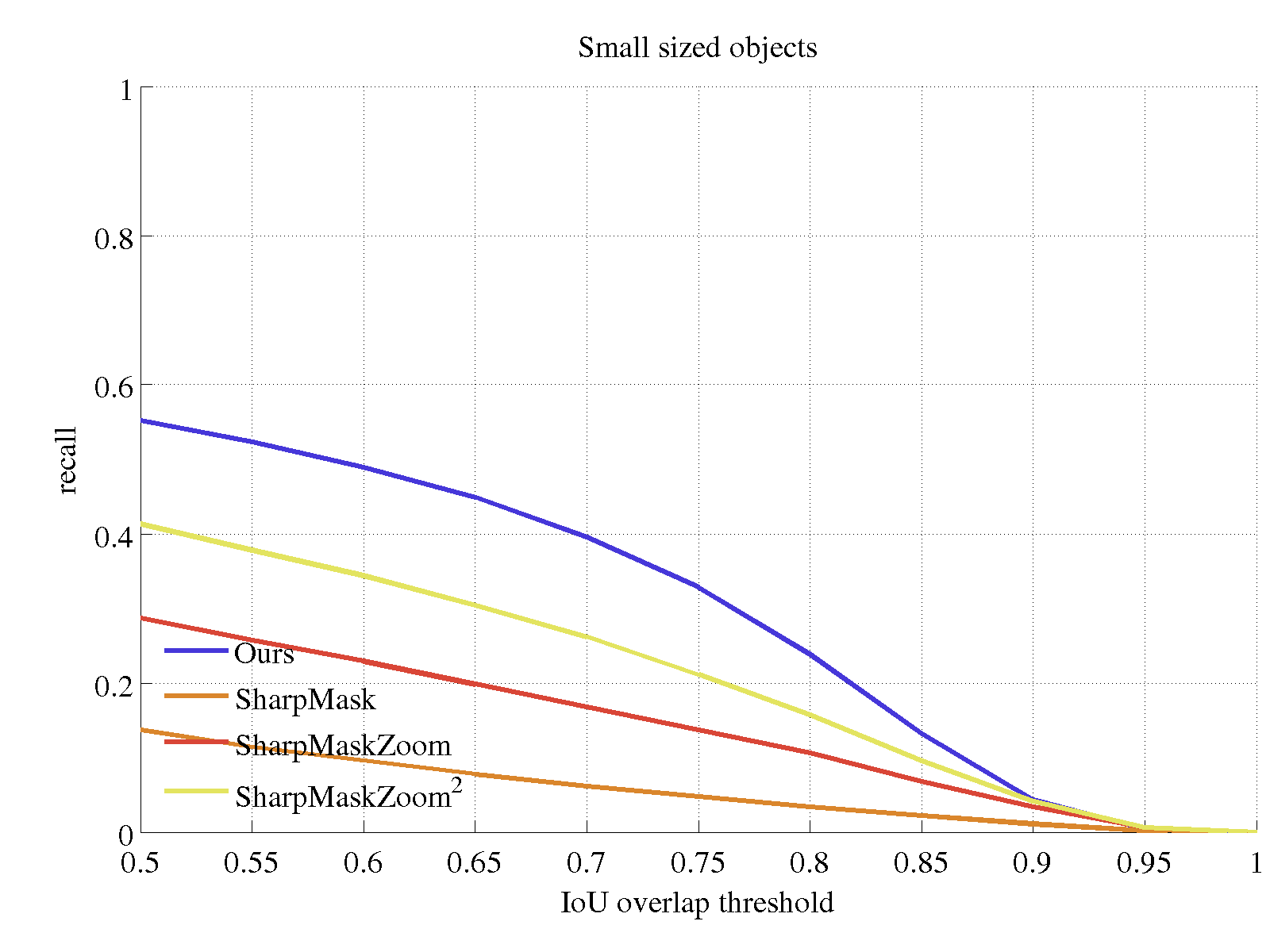}}&
\bmvaHangBox{\includegraphics[width=4.0cm]{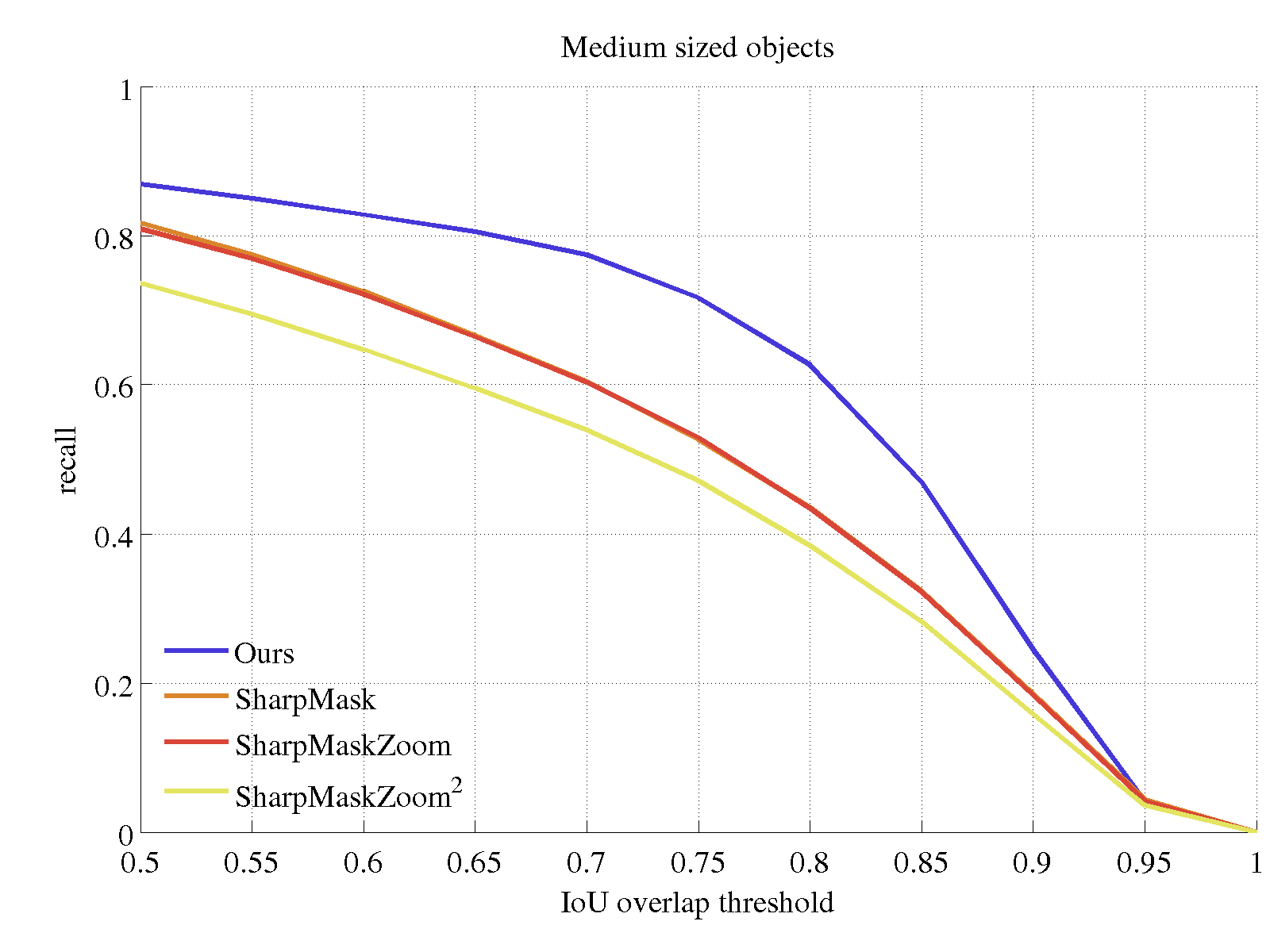}}&
\bmvaHangBox{\includegraphics[width=4.0cm]{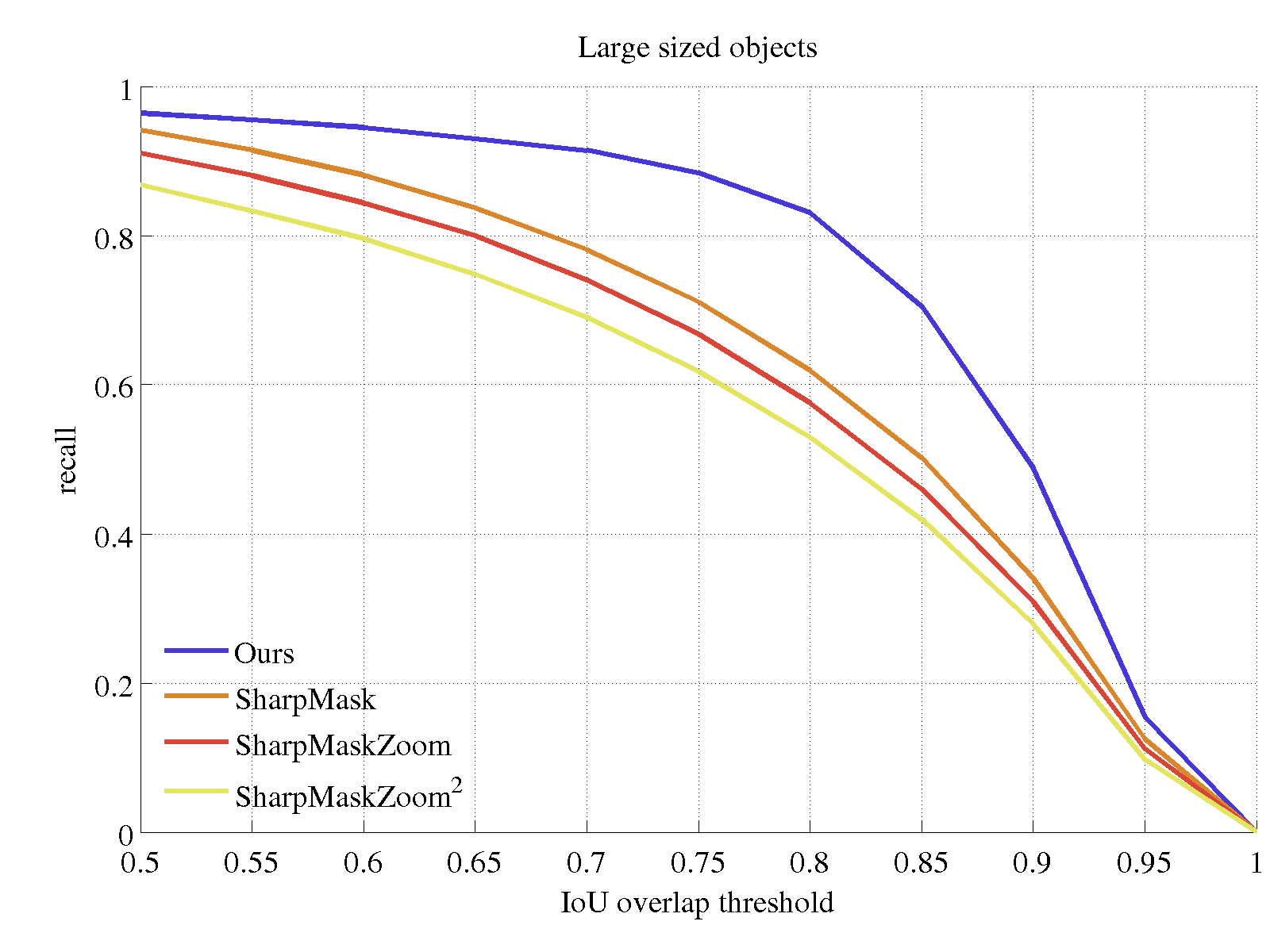}}\\
(d)&(e)&(f)
\end{tabular}}}
\resizebox{\textwidth}{!}{
{\setlength{\extrarowheight}{8pt}\scriptsize
\begin{tabular}{cccc}
\bmvaHangBox{\includegraphics[width=5cm]{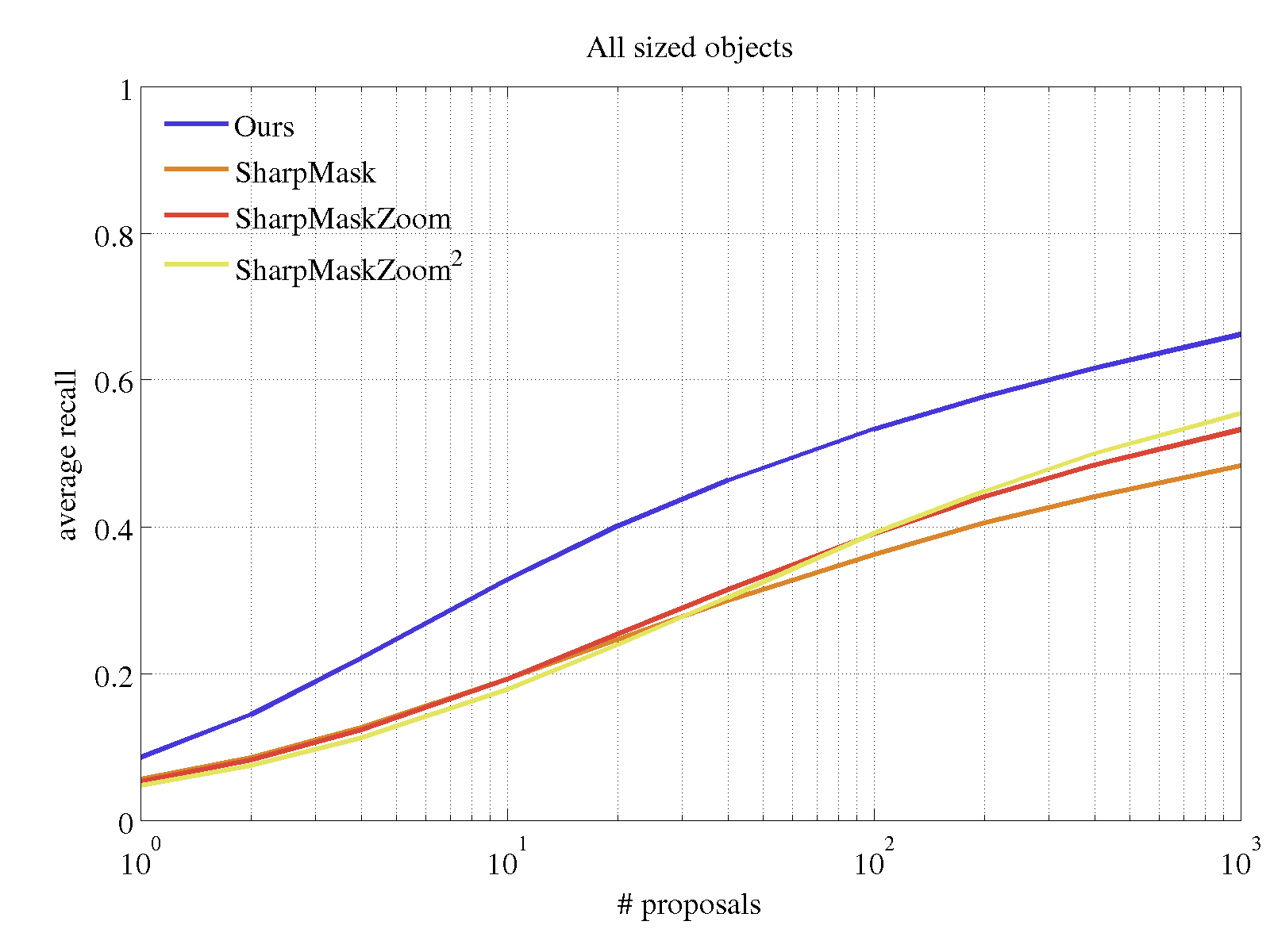}}&\hspace{-15pt}
\bmvaHangBox{\includegraphics[width=5cm]{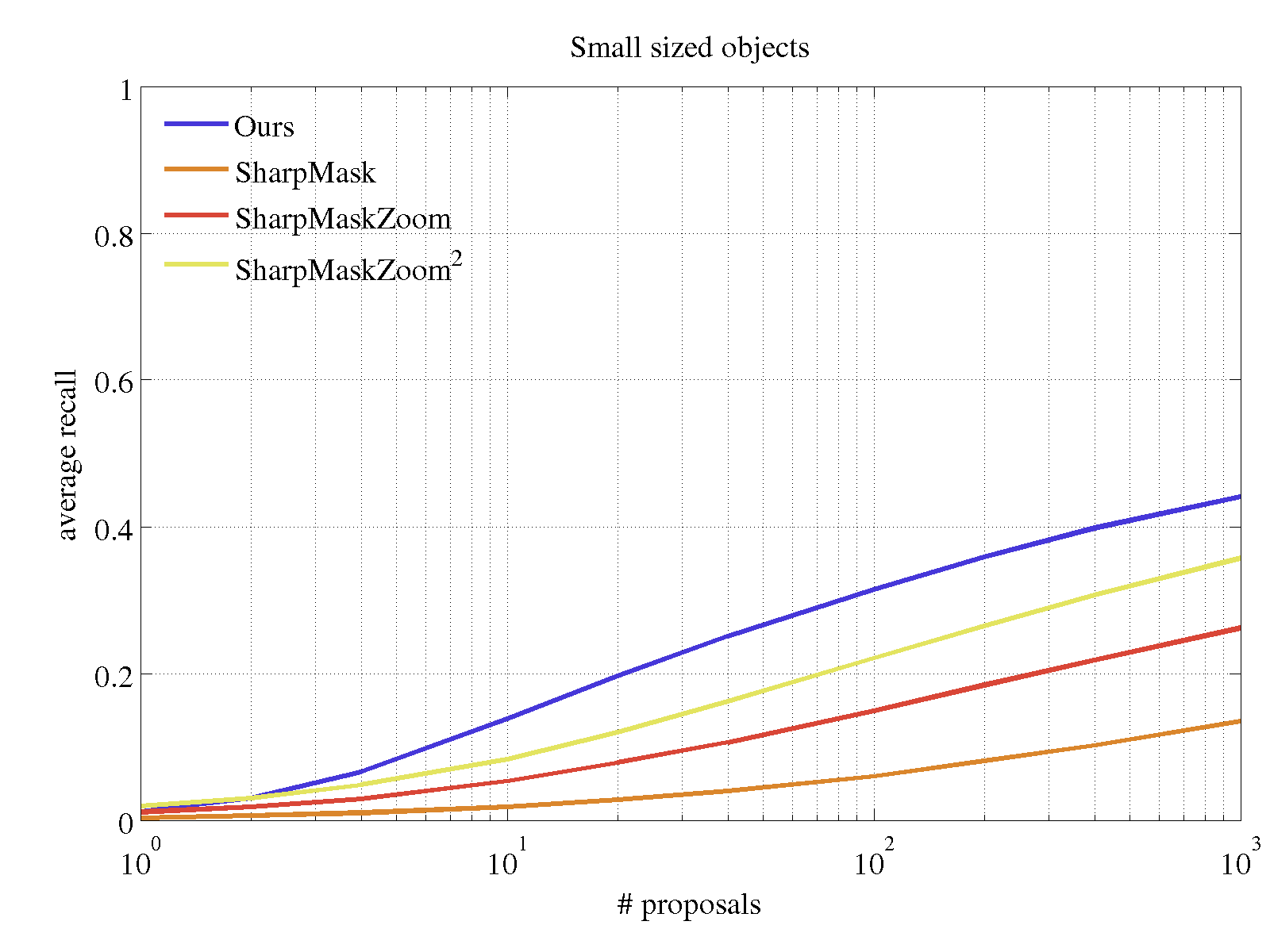}}&
\hspace{-15pt}
\bmvaHangBox{\includegraphics[width=5cm]{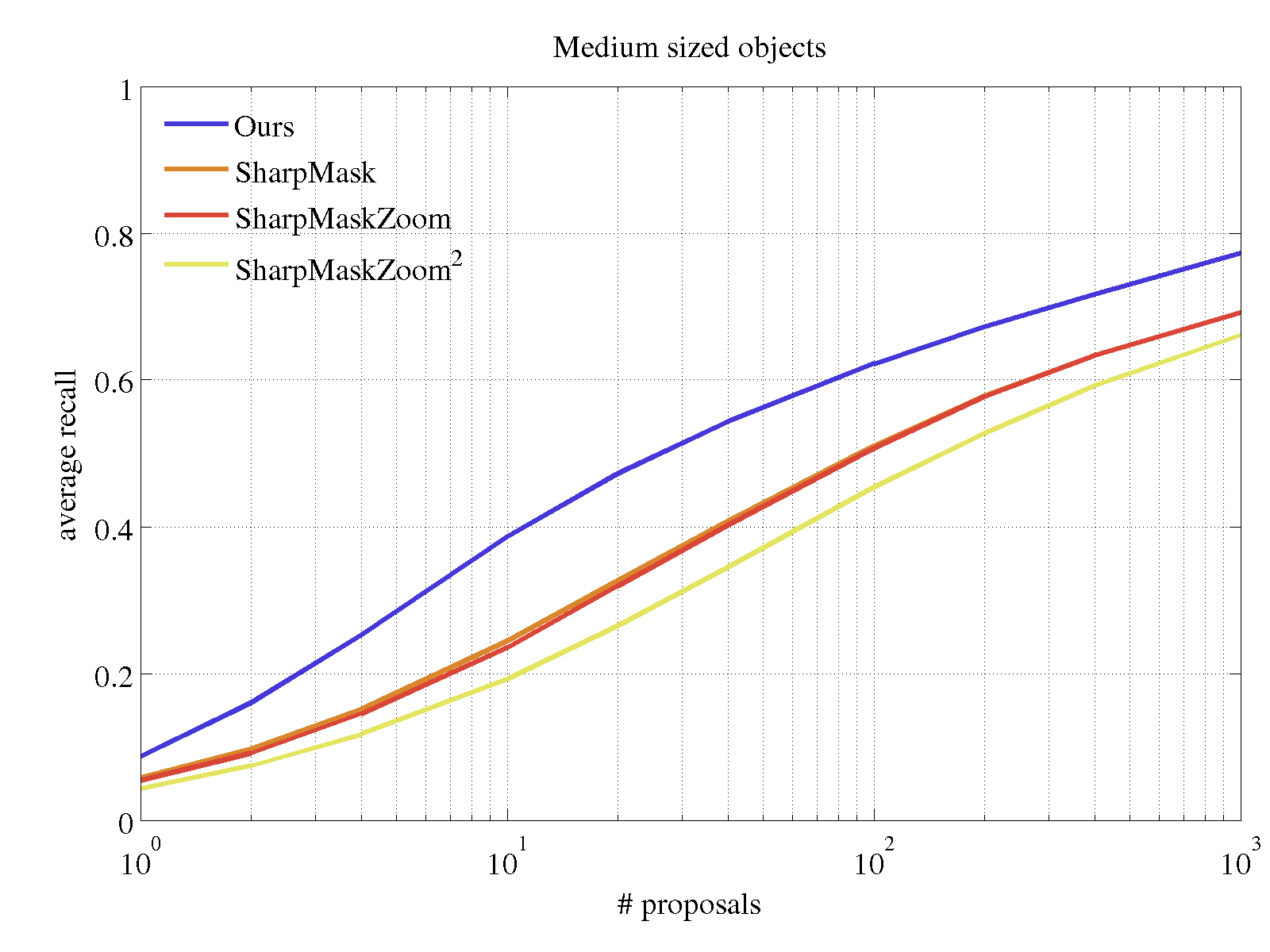}}&
\hspace{-15pt}
\bmvaHangBox{\includegraphics[width=5cm]{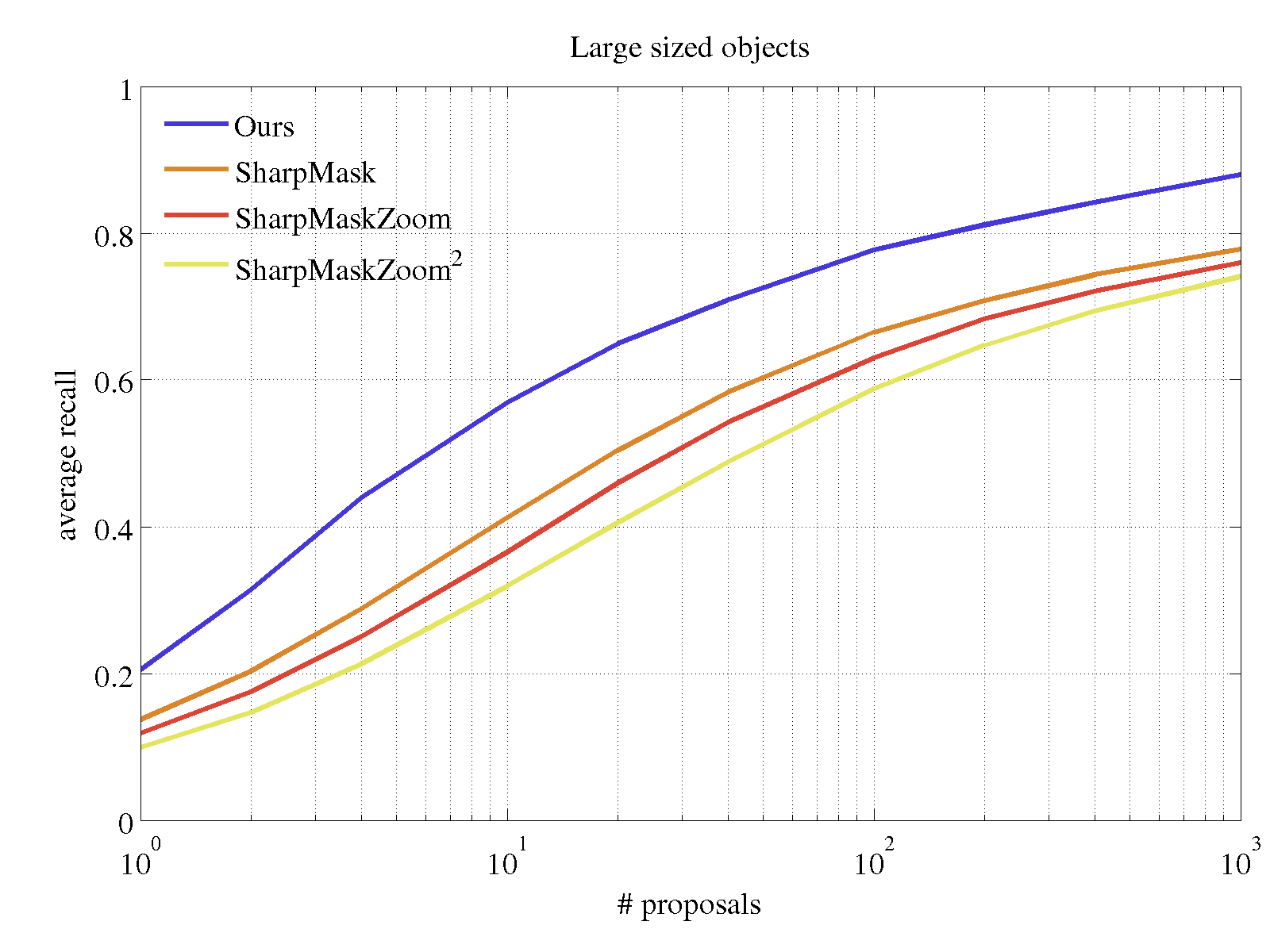}}\\
\large{(g)}&\large{(h)}&\large{(i)}&\large{(j)}
\end{tabular}}}
\end{center}
\vspace{-6pt}
\caption{
\small{\textbf{Comparison with previous state-of-the-art.}
Comparison of our \emph{AttractioNet} box proposal model (\emph{Ours} entry) against the previous state-of-the-art~\cite{PinheiroLCD16} (\emph{SharpMask}, \emph{SharpMaskZoom} and \emph{SharpMaskZoom$^2$} entries) w.r.t. the recall versus IoU trade-off and average recall versus proposals number trade-off that they achieve under various test scenarios.
Specifically, the sub-figures (a), (b) and (c) plot the recall as a function of the IoU threshold for 10, 100 and 1000 box proposals respectively and 
the sub-figures (d), (e) and (f) plot the recall as a function of the IoU threshold for 100 box proposals and with respect to the small, medium and large sized objects correspondingly. 
Also, the sub-figures (g), (h), (i) and (j) plot the average recall as a function of the proposals number for all the objects regardless of their size as well as for the small, medium and large sized objects respectively.
The reported results are from the first $5k$ images of the COCO validation set.}}
\label{fig:comparison}
\vspace{-10pt}
\end{figure}

\textbf{Comparison with previous state-of-the-art.} In Figure~\ref{fig:comparison} we compare the box proposals generated from our \emph{AttractioNet} model (\emph{Ours} entry) against those generated from the previous state-of-the-art~\cite{PinheiroLCD16} (entries \emph{SharpMask}, \emph{SharpMaskZoom} and \emph{SharpMaskZoom}$^2$) w.r.t. the recall versus IoU trade-off and average recall versus proposals number trade-off that they achieve. 
Also, in Table~\ref{tab:AR_COCO} we report the AR results both for our method and for the SharpMask entries. 
We observe that the model proposed in our work has clearly superior performance over the SharpMask entries under all test cases.

\vspace{-10pt}

\subsubsection{Ablation study}
\begin{figure}[t!]
\begin{center}
\bmvaHangBox{\includegraphics[width=7cm]{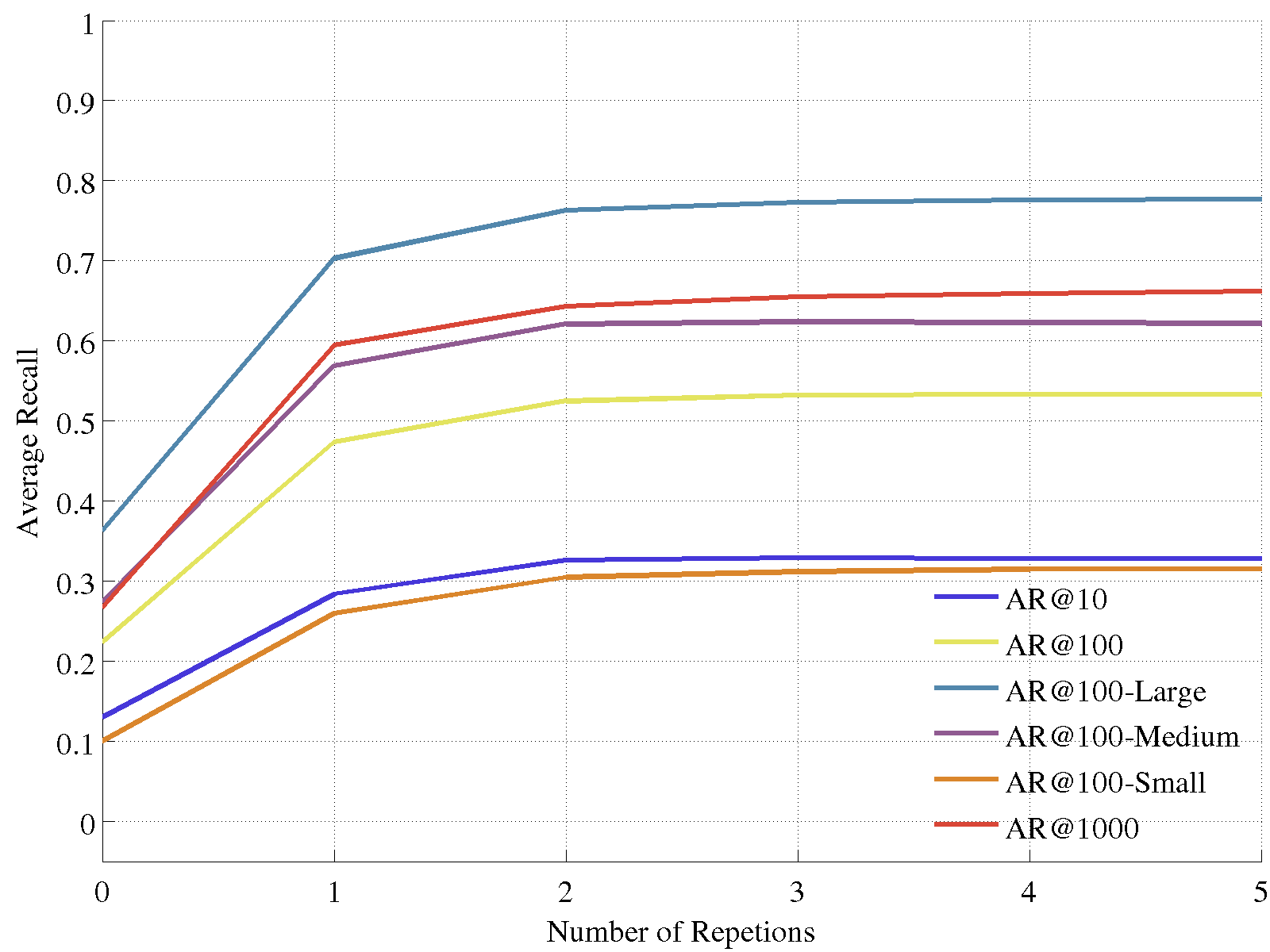}}\\
\end{center}
\vspace{-8pt}
\caption{\small{Average recall versus the repetitions number of the active box proposal generation algorithm in the COCO validation set. Note that 0 repetitions is the scenario of simply applying the objectness module on the seed boxes.}}
\label{fig:ar_iter}
\vspace{-10pt}
\end{figure}
\begin{table}[t!]
\centering
\resizebox{\textwidth}{!}{
{\setlength{\extrarowheight}{2pt}\scriptsize
\begin{tabular}{c c | c | c c c c c c c  }
\hline
Box refinement & Active box generation & \# attended boxes & AR@10 & AR@100 & AR@1000 & AR@100-Small & AR@100-Medium & AR@100-Large\\
\hline
&  & $18k$ & 0.147 & 0.260 & 0.326 & 0.122 & 0.317 & 0.412 \\
\checkmark &  & $18k$ & 0.298 & 0.491 & 0.622 & 0.281 & 0.583 & 0.717 \\
\checkmark & \checkmark & $18k$ &\textbf{0.328} & \textbf{0.533} & \textbf{0.662} & \textbf{0.315} & \textbf{0.622} & \textbf{0.777} \\
\hline
\end{tabular}}}
\vspace{-8pt}
\caption{
\small{\textbf{Ablation study of our \emph{AttractioNet} box proposal system.} 
In the first row we simply apply the objectness scoring module on a set of $18k$ seed boxes.
In the second row we apply on the same set of $18k$ seed boxes both the objectness scoring module and the box refinement module.
In the last row we utilize our full active box generation strategy that in total attends $18k$ boxes of which $10k$ are seed boxes and the rest $8k$ boxes are actively generated.
The reported results are from the first $5k$ images of COCO validation set. }}
\label{tab:AR_COCO_ACT}
\vspace{-10pt}
\end{table}

Here we perform an ablation study of our two key ideas for improving the state-of-the-art on the bounding box proposal generation task:\\

\textbf{Object location refinement module.} 
In order to assess the importance of our object location refinement module we evaluated two test cases for generating box proposals: 
\textbf{(1)}
simply applying the objectness scoring module on a set of $18k$ seed boxes (first row of Table~\ref{tab:AR_COCO_ACT}) and \textbf{(2)} applying both the objectness scoring module and the object location refinement module on the same set of $18k$ seed boxes (second row of Table~\ref{tab:AR_COCO_ACT}). 
Note that in none of them is the active box generation strategy being used. 
The average recall results of those two test cases are reported in the first two rows of Table~\ref{tab:AR_COCO_ACT}. 
We observe that without the object location refinement module the average recall performance of the box proposal system is very poor. 
In contrast, the average recall performance of the test case that involves the object location refinement module but not the active box generation strategy is already better than the previous state-of-the-art as reported in Table~\ref{tab:AR_COCO}, which demonstrates the very good localization accuracy of our category agnostic location refinement module.\\

\textbf{Active box generation strategy.} 
Our active box generation strategy, which we call \emph{Attend Refine Repeat} algorithm, attends in total $18k$ boxes before it outputs the final list of box proposals.
Specifically, it attends $10k$ seed boxes in the first repetition of the algorithm and $2k$ actively generated boxes in each of the following four repetitions.
A crucial question is whether actively generating those extra $8k$ boxes is really essential in the task or we could achieve the same average recall performance by directly attending $18k$ seed boxes and without continuing on the active box generation stage. 
We evaluated such test case and we report the average recall results in Table~\ref{tab:AR_COCO_ACT} (see rows 2 and 3). 
We observe that employing the active box generation strategy (3rd row in Table~\ref{tab:AR_COCO_ACT}) offers a significant boost in the average recall performance (between 3 and 6 absolute points in the percentage scale) thus proving its importance on yielding well localized bounding box proposals.
Also, in the right side of Figure~\ref{fig:ar_iter} we plot the average recall metrics as a function of the repetitions number of our active box generation strategy. 
We observe that the average recall measurements are increased as we increase the repetitions number and that the increase is more steep on the first repetitions of the algorithm while it starts to converge after the 4th repetition.
\vspace{-10pt}

\subsubsection{Run time} \label{sec:run_time}

In the current work we did not focus on providing an optimized implementation of our approach.
There is room for significantly improving computational efficiency.
For instance, just by using SVD decomposition on the fully connected layers of the objectness module at post-training time (similar to Fast-RCNN~\cite{girshick2015fast}) and early stopping a sequence of bounding box location refinements in the case it has already converged\footnote{
A sequence of bounding box refinements is considered that it has converged when the IoU between the two lastly predicted boxes in the sequence is greater than 0.9.}, the runtime drops from 4.0 seconds to 1.63 seconds without losing almost no accuracy (see Table~\ref{tab:AR_COCO_RUN_TIME}).
There are also several other possibilities that we have not yet explored such as tuning the number of feature channels and/or network layers of the CNN architecture (similar to the DeepBox~\cite{kuo2015deepbox} and the SharpMask~\cite{PinheiroLCD16} approaches). 

In the remainder of this section we will use the fast version of our \emph{AttractioNet} approach in order to provide experimental results.

\begin{table}[t!]
\centering
\resizebox{\textwidth}{!}{
{\setlength{\extrarowheight}{2pt}\scriptsize
\begin{tabular}{l | c | c c c c c c }
\hline
Method & Run time & AR@10 & AR@100 & AR@1000 & AR@100-Small & AR@100-Medium & AR@100-Large\\
\hline
\multicolumn{8}{c}{COCO validation set}\\
\hline
AttractioNet (Ours) &  4.00 sec &\textbf{0.328} & \textbf{0.533} & \textbf{0.662} & 0.315 & \textbf{0.622} & \textbf{0.777} \\
AttractioNet (Ours, fast version) &  1.63 sec & 0.326 & 0.532 & 0.660 & \textbf{0.317} & 0.621 & 0.771 \\
\hline
\multicolumn{8}{c}{VOC2007 test set}\\
\hline
AttractioNet (Ours) & 4.00 sec & \textbf{0.554} & \textbf{0.744} & \textbf{0.859} & 0.562 & \textbf{0.670} & \textbf{0.794}\\
AttractioNet (Ours, fast version) & 1.63 sec & 0.547 & 0.740 & 0.848 & \textbf{0.575} & 0.666 & 0.788 \\
\hline
\end{tabular}}}
\vspace{-8pt}
\caption{
\small{\textbf{Run time of our approach on a GTX Titan X GPU.} 
The reported results are from the first $5k$ images of COCO validation set and the PASCAL VOC2007 test set.}}
\label{tab:AR_COCO_RUN_TIME}
\vspace{-10pt}
\end{table}

\subsection{Generalization to unseen categories} \label{sec:generalization}

So far we have evaluated our \emph{AttractioNet} approach 
--- in the end task of object box proposal generation ---
on the COCO validation set and the PASCAL VOC2007 test set that are labelled with the same or a subset of the object categories "seen" in the training set.
In order to assess the \emph{AttractioNet}'s capability to generalize to "unseen" categories, as it is suggested by Chavali \etal~\cite{chavali2015object}, we evaluate our \emph{AttractioNet} model on two extra datasets that are labelled with object categories that are not present in its training set ("unseen" object categories).\\

\textbf{From COCO to ImageNet~\cite{russakovsky2014imagenet}.}
Here we evaluate our COCO trained \emph{AttractioNet} box proposal model on the ImageNet~\cite{russakovsky2014imagenet} ILSVRC2013 detection task validation set that is labelled with 200 different object categories and we report average recall results in Table~\ref{tab:AR_GENERALIZATION2}.
Note that among the 200 categories of ImageNet detection task, 60 of them, as we identified, are also present in the \emph{AttractioNet}'s training set (see Appendix~\ref{sec:imagenet_coco}).
Thus, for a better insight on the generalization capabilities of \emph{AttractioNet},
we divided the ImageNet detection task categories on two groups, the "seen" by \emph{AttractioNet} categories and the "unseen" categories, and we report the average recall results separately for those two groups of object categories in Table~\ref{tab:AR_GENERALIZATION2}.
For comparison purposes we also report the average recall performance of a few indicative other box proposal methods that their code is publicly available. 
We observe that, despite the performance difference of our approach between the "seen" and the "unseen" object categories (which is to be expected), its average recall performance on the "unseen" categories is still quite high and significantly better than the other box proposal methods.
Note that even the non-learning based approaches of Selective Search and EdgeBoxes exhibit a performance drop on the "unseen" by \emph{AttractioNet} group of object categories, which we assume is because this group contains more intrinsically difficult to discover objects.\\

\begin{table}[t!]
\centering
\resizebox{\textwidth}{!}{
{\setlength{\extrarowheight}{2pt}\scriptsize
\begin{tabular}{l | c c c | c c c | c c c   }
\hline
\multirow{2}{*}{Method} & \multicolumn{3}{c|}{All categories} & \multicolumn{3}{c|}{Seen categories} & \multicolumn{3}{c}{Unseen categories}\\
 & AR@10 & AR@100 & AR@1000 & AR@10 & AR@100 & AR@1000 & AR@10 & AR@100 & AR@1000\\
\hline
AttractioNet (Ours) & \textbf{0.412} & \textbf{0.618} & \textbf{0.748} & \textbf{0.474} & \textbf{0.671} & \textbf{0.789} & \textbf{0.299} & \textbf{0.521} & \textbf{0.673} \\
\hline
EdgeBoxes~\cite{zitnick2014edge}                & 0.182 & 0.377 & 0.550 & 0.194 & 0.396 & 0.566 & 0.160 & 0.344 & 0.519\\
Selective Search~\cite{van2011segmentation} & 0.132 & 0.358 & 0.562 & 0.143 & 0.372 & 0.568 & 0.111 & 0.332 & 0.551\\
MCG~\cite{APBMM2014}                                    & 0.219 & 0.428 & 0.603 & 0.228 & 0.447 & 0.623 & 0.205 & 0.395 & 0.568\\
\hline
\end{tabular}}}
\vspace{-8pt}
\caption{
\small{\textbf{Generalization to unseen categories: from COCO to ImageNet.} In this table we report average recall results on the ImageNet~\cite{russakovsky2014imagenet} ILSVRC2013 detection task validation set that includes around $20k$ images and it is labelled with 200 object categories. \textbf{\emph{Seen categories}} are the set of object categories that our COCO trained \emph{AttractioNet} model "saw" during training. In contrast, \textbf{\emph{unseen categories}} is the set of object categories that were not present in the training set of our \emph{AttractioNet} model.}}
\label{tab:AR_GENERALIZATION2}
\vspace{-10pt}
\end{table}

\begin{table}[t!]
\centering
\resizebox{\textwidth}{!}{
{\setlength{\extrarowheight}{2pt}\scriptsize
\begin{tabular}{l | c c c c c c }
\hline
Method & AR@10 & AR@100 & AR@1000 & AR@100-Small & AR@100-Medium & AR@100-Large\\
\hline
AttractioNet (Ours) &\textbf{0.159} & \textbf{0.389} & \textbf{0.579} & \textbf{0.205} & \textbf{0.419} & \textbf{0.498} \\
\hline
EdgeBoxes~\cite{zitnick2014edge} & 0.049 & 0.160 & 0.362 & 0.020 & 0.131 & 0.332 \\
Selective Search~\cite{van2011segmentation} &0.024 & 0.143 & 0.422 & 0.008 & 0.085 & 0.362 \\
MCG~\cite{APBMM2014} & 0.078 & 0.237 & 0.441 & 0.045 & 0.195 & 0.476 \\
\hline
\end{tabular}}}
\vspace{-8pt}
\caption{
\small{\textbf{Generalization to unseen categories: from COCO to NYU-Depth V2 dataset.}
In this table we report average recall results on the 1449 labelled images of the NYU-Depth V2 dataset~\cite{SilbermanECCV12}. Note that the NYU-Depth V2 dataset is densely labelled with more than 800 different categories.}}
\label{tab:AR_GENERALIZATION3}
\vspace{-10pt}
\end{table}

\textbf{From COCO to NYU Depth dataset~\cite{SilbermanECCV12}.}
The NYU Depth V2 dataset~\cite{SilbermanECCV12} provides 1449 images (recorded from indoor scenes)
that are densely pixel-wise annotated with 864 different categories. 
We used the available instance-wise segmentations to create ground truth bounding boxes and we tested our COCO trained \emph{AttractioNet} model on them (see Table~\ref{tab:AR_GENERALIZATION3}).
Note that among the 864 available pixel categories, a few of them are "stuff" categories (e.g. wall, floor, ceiling or stairs) or in general non-object pixel categories that our object box proposal method should by definition not recall.
Thus, during the process of creating the ground truth bounding boxes, those non-object pixel segmentation annotations were excluded (see Appendix~\ref{sec:ignored_categories}).
In Table~\ref{tab:AR_GENERALIZATION3} we report the average recall results of our \emph{AttractioNet} method as well as of a few other indicative methods that their code is publicly available. 
We again observe that  our method surpasses  all other approaches by a significant margin.
Furthermore, in this case the superiority of our approach is more evident on the average recall of the small and medium sized objects.\\

To conclude we argue that our learning based \emph{AttractioNet} approach exhibits good generalization behaviour.
Specifically, 
its average recall performance on the "unseen" object categories remains very high and is also much better than other competing approaches, including both  learning-based approaches such as the MSG  and hand-engineered ones such as the Selective Search or the EdgeBoxes methods. 
A performance drop is still observed while going from "seen" to "unseen" categories, but this is something to be expected   given that any machine learning  algorithm will always exhibit a certain performance drop while going from "seen" to "unseen" data (i.e. training set accuracy versus test set accuracy).

\vspace{-10pt}

\subsection{AttractioNet box proposals evaluation in the context of the object detection task} \label{sec:detection}

Here we evaluate our \emph{AttractioNet} box proposals in the context of the object detection task by training and testing a box proposal based object detection system on them (specifically we use the fast version of \emph{AttractioNet} that is described in section~\ref{sec:run_time}).\\

\textbf{Detection system.}
Our box proposal based object detection system consists of a Fast-RCNN~\cite{girshick2015fast} category-specific recognition module and a LocNet Combined ML~\cite{gidaris2016locnet} category-specific bounding box refinement module (see Appendix~\ref{sec:detection_system} for more details).
As post-processing we use a non-max-suppression step (with IoU threshold of 0.35) that is enhanced with the box voting technique described in the MR-CNN system~\cite{gidaris2015object} (with IoU threshold of 0.75).
Note that we did not include iterative object localization as in the LocNet~\cite{gidaris2016locnet} or MR-CNN~\cite{gidaris2015object} papers, since our bounding box proposals are already very well localized and we did not get any significant improvement from running the detection system for extra iterations. 
Using the same trained model we provide results for two test cases: 
\textbf{(1)} using a single scale of 600 pixels during test time and \textbf{(2)} using two scales of 500 and 1000 pixels during test time.\\

\textbf{Detection evaluation setting.}
The detection evaluation metrics that we use are the average precision (AP) for the IoU thresholds of $0.50$ (AP$@0.50$), $0.75$ (AP$@0.75$) and the COCO style of average precision (AP$@0.50:0.95$) that averages the traditional AP over several IoU thresholds between $0.50$ and $0.95$.
Also, we report the COCO style of average precision with respect to the small (AP$@$Small), medium (AP$@$Medium) and large (AP$@$Large) sized objects. 
We perform the evaluation on $5k$ images of COCO 2014 validation set and we provide final results on the COCO 2015 test-dev set.\\

\begin{figure}[t!]
\begin{center}
\resizebox{\textwidth}{!}{
{\setlength{\extrarowheight}{2pt}\scriptsize
\begin{tabular}{cc}
\bmvaHangBox{\includegraphics[width=5.0cm]{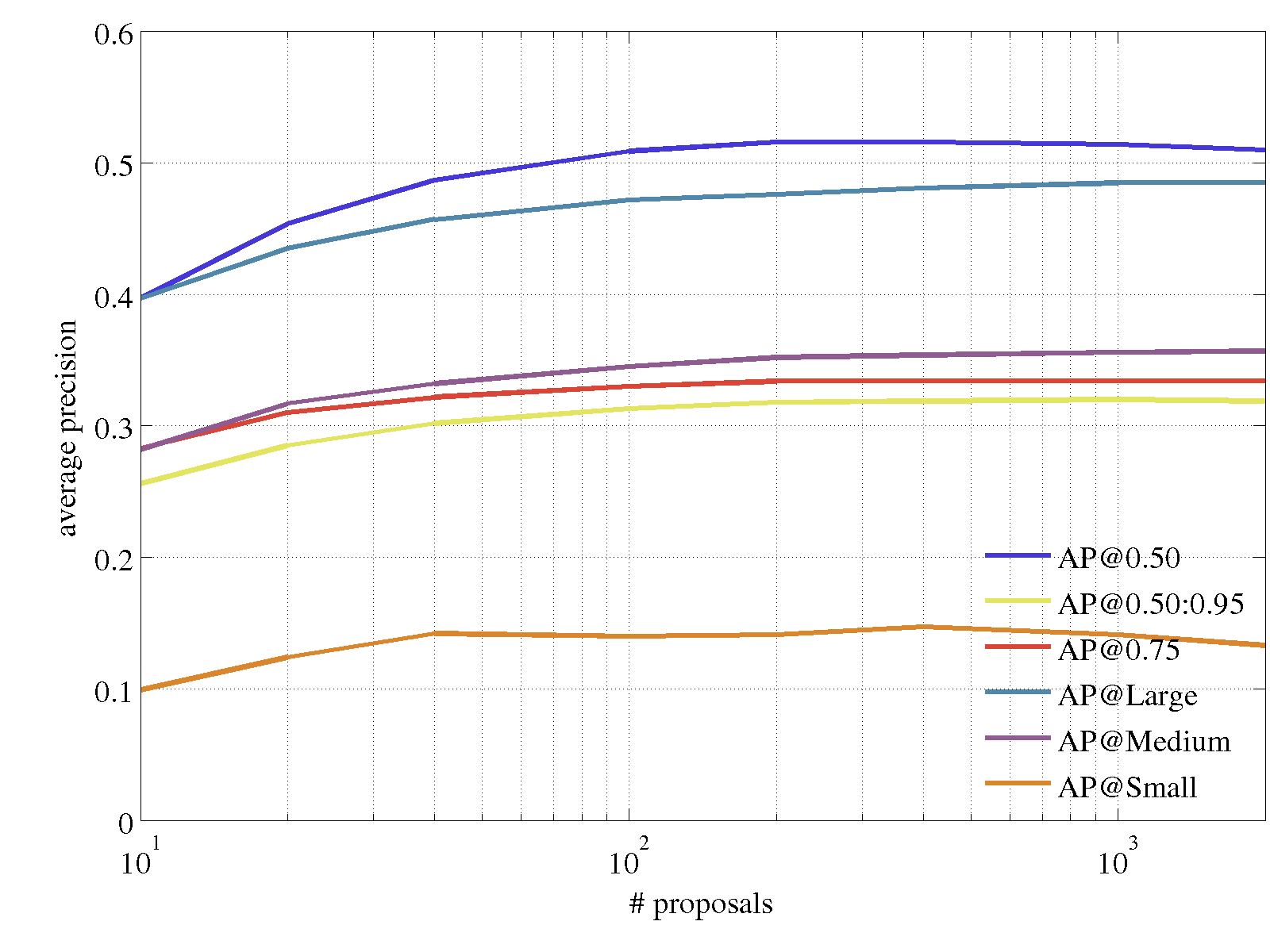}}&
\bmvaHangBox{\includegraphics[width=5.0cm]{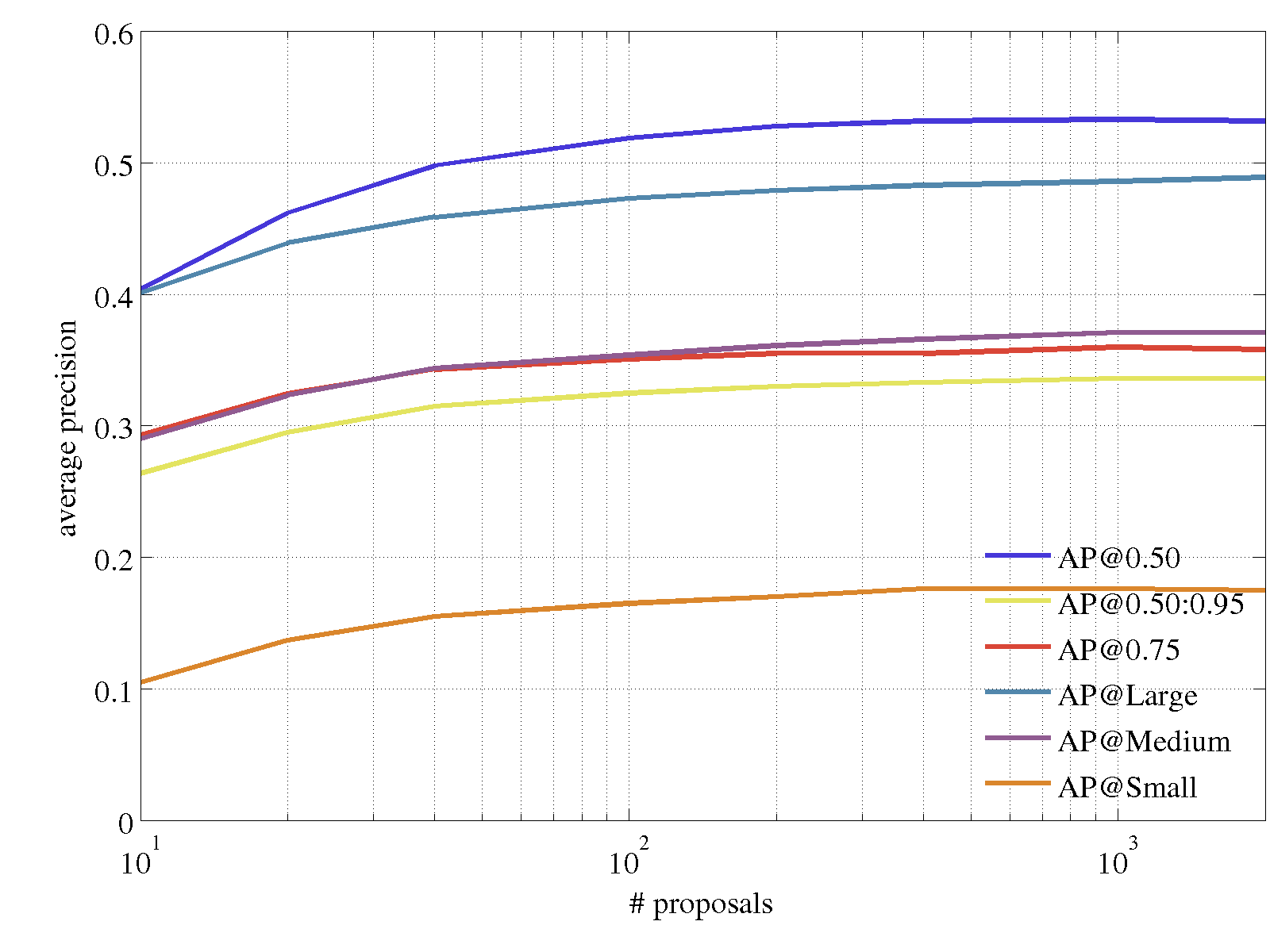}}\\
(a)&(b)\\
\end{tabular}}}
\end{center}
\vspace{-10pt}
\caption{
\small{\textbf{Detection results: Average precision versus \emph{AttractioNet} box proposals number.} 
\textbf{(a)} During test time a single scale of 600 pixels is being used.
\textbf{(b)} During test time two scales of 500 and 1000 pixels are being used.
The reported results are from $5k$ images of COCO validation set.}}
\label{fig:detection}
\vspace{-10pt}
\end{figure}

\textbf{Detection results.}
In Figure~\ref{fig:detection} we provide plots of the achieved average precision (AP) as a function of the used box proposals number and in Table~\ref{tab:coco_detection} we provide the average precision results for 10, 100, 1000 and 2000 box proposals.
We observe that in all cases, the average precision performance of the detection system seems to converge after the 200 box proposals.
Furthermore, for single scale test case our best COCO-style average precision is 0.320 and for the two scales test case our best COCO-style average precision is 0.337.
By including horizontal image flipping augmentation during test time our COCO-style average precision performance is increased to 0.343. 
Finally, in Table~\ref{tab:coco_detection_test} we provide the average precision performance in the COCO test-dev 2015 set where we achieve a COCO-style AP of 0.341. By comparing with the average precision performance of the other competing methods, we observe that:
\begin{compactitem}
\item Comparing with the other VGG16-Net based object detection systems (ION~\cite{bell2015inside} and MultiPath~\cite{zagoruyko2016multipath} systems), our detection system achieves the highest COCO-style average precision with its main novelties w.r.t. the Fast R-CNN~\cite{girshick2015fast} baseline being \textbf{(1)} the use of the \emph{AttractioNet} box proposals that are introduced in this paper and \textbf{(2)} the LocNet~\cite{gidaris2016locnet} category specific object location refinement technique that replaces the bounding box regression step. 
\item Comparing with the ION~\cite{bell2015inside} detection system, which is also VGG16-Net based, our approach is better on the COCO-style AP metric (that favours good object localization) while theirs is better on the typical AP$@$0.50 metric. We hypothesize that this is due to the fact that our approach targets to mainly improve the localization aspect of object detection by improving the box proposal generation step while theirs the recognition aspect of object detection.
The above observation suggests that many of the novelties introduced on the ION~\cite{bell2015inside} and MultiPath~\cite{zagoruyko2016multipath} systems w.r.t. object detection could be orthogonal to our box proposal generation work.
\item The achieved average precision performance of our VGG16-Net based detection system is close to the state-of-the-art \emph{ResNet-101 based} Faster R-CNN+++ detection system~\cite{he2015deep} 
that exploits the recent successes in deep representation learning introduced --- under the name Deep Residual Networks --- in the same work by He \etal~\cite{he2015deep}. 
Presumably, our overall detection system could also be benefited by being based on the Deep Residual Networks~\cite{he2015deep} or the more recent wider variant called Wide Residual Networks~\cite{ZagoruykoK16}. 
\item Finally, our detection system has the highest average precision performance w.r.t. the small sized objects, which is a challenging problem, surpassing by a healthy margin even the ResNet-101 based Faster R-CNN+++ detection system~\cite{he2015deep}. 
This is thanks to the high average recall performance of our box proposal method on the small sized objects.
\end{compactitem}

\begin{table}
\begin{center}
\resizebox{\textwidth}{!}{
{\setlength{\extrarowheight}{2pt}\scriptsize
{\begin{tabular}{c|r||cccccc}
\hline
Test scale(s) & \# proposals & AP$@$0.50 & AP$@$0.75 & AP$@$0.50:0.95 & AP$@$Small & AP$@$Medium & AP$@$Large\\
\hline
$600px$ & 10   & 0.397 & 0.283 & 0.256 & 0.099 & 0.282 & 0.397\\
$600px$ & 100  & 0.509 & 0.330 & 0.313 & 0.140 & 0.345 & 0.472\\
$600px$ & 1000 & 0.514 & 0.334 & 0.320 & 0.141 & 0.356 & 0.485\\
$600px$ & 2000 & 0.510 & 0.334 & 0.319 & 0.133 & 0.357 & 0.485\\
\hline
$500px,1000px$ & 10   & 0.404 & 0.293 & 0.264 & 0.105 & 0.290 & 0.401\\
$500px,1000px$ & 100  & 0.519 & 0.351 & 0.325 & 0.165 & 0.354 & 0.473\\
$500px,1000px$ & 1000 & 0.533 & 0.360 & 0.336 & 0.176 & 0.371 & 0.486\\
$500px,1000px$ & 2000 & 0.532 & 0.358 & 0.336 & 0.175 & 0.371 & 0.489\\
\hline
$500px,1000px$ $\bigstar$ & 2000 & \textbf{0.540} & \textbf{0.364} & \textbf{0.343} & \textbf{0.184} & \textbf{0.382} & \textbf{0.491}\\
\hline
\end{tabular}}}}
\end{center}
\caption{\small{\textbf{Detection results: Average precision performance using \emph{AttractioNet} box proposals.} 
The reported results are from $5k$ images of COCO validation set. The last entry with the $\bigstar$ symbol uses horizontal image flipping augmentation during test time.}}
\label{tab:coco_detection}
\vspace{-15pt}
\end{table}

\begin{table}
\begin{center}
\resizebox{\textwidth}{!}{
{\setlength{\extrarowheight}{2pt}\scriptsize
{\begin{tabular}{l | l||cccccc}
\hline
Method & Base CNN & AP$@$0.50 & AP$@$0.75 & AP$@$0.50:0.95 & AP$@$Small & AP$@$Medium & AP$@$Large\\
\hline
\emph{AttractioNet} based detection system (Ours) & VGG16-Net~\cite{simonyan2014very} & 0.537 & 0.363 & 0.341 & 0.175 & 0.365 & 0.469\\
\hline
ION~\cite{bell2015inside} & VGG16-Net~\cite{simonyan2014very} & 0.557 & 0.346 & 0.331 & 0.145 & 0.352 & 0.472\\
MultiPath~\cite{zagoruyko2016multipath} & VGG16-Net~\cite{simonyan2014very} &   - & - & 0.315 & - & - & -\\
Faster R-CNN+++~\cite{he2015deep} & ResNet-101~\cite{he2015deep} & 0.557 & - & 0.349 & 0.156 & 0.387 & 0.509\\
\hline
\end{tabular}}}}
\end{center}
\caption{\small{\textbf{Detection results in COCO test-dev 2015 set.} 
In this table we report the average precision performance of our \emph{AttractioNet} box proposals based detection system that uses 2000 proposals and two test scales of 500 and 1000 pixels. Note that: (1) all methods in this table (including ours) use horizontal image flipping augmentation during test time, (2) the ION~\cite{bell2015inside} and MultiPath~\cite{zagoruyko2016multipath} detection systems use a single test scale of 600 and 800 pixels respectively while the Faster R-CNN+++ entry uses the scales \{200, 400, 600, 800, 1000\}, (3) apart from the ResNet-101 based Faster R-CNN+++~\cite{he2015deep} entry, all the other methods are based on the VGG16-Network~\cite{simonyan2014very}, (4) the reported results of all the competing methods are from the single model versions of their systems (and not the model ensemble versions) and (5) the reported results of the MultPath system are coming from $5k$ images of the COCO validation set (however, we expect the AR metrics on the test-dev set to be roughly similar). }}
\label{tab:coco_detection_test}
\vspace{-15pt}
\end{table}

\subsection{Qualitative results} \label{sec:qualitative}
\begin{figure}
\begin{center}
\bmvaHangBox{\includegraphics[width=3cm]{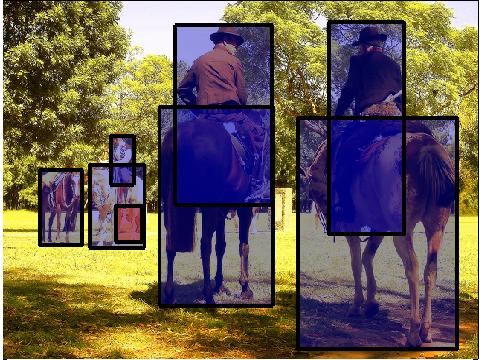}
\includegraphics[width=3cm]{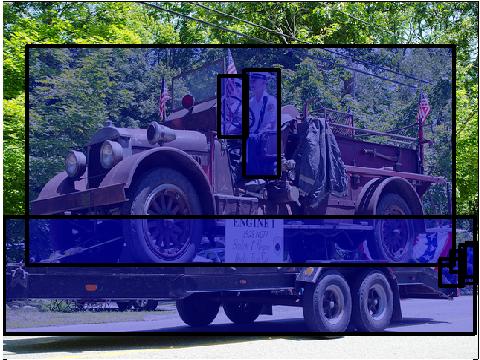}
\includegraphics[width=3cm]{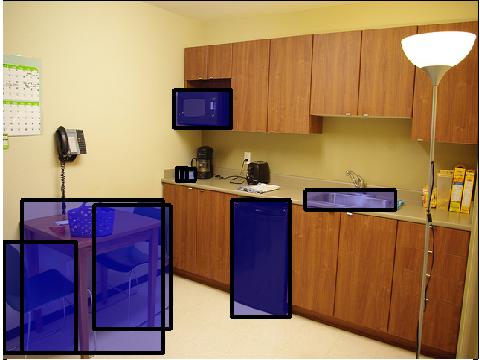}
\includegraphics[width=3cm]{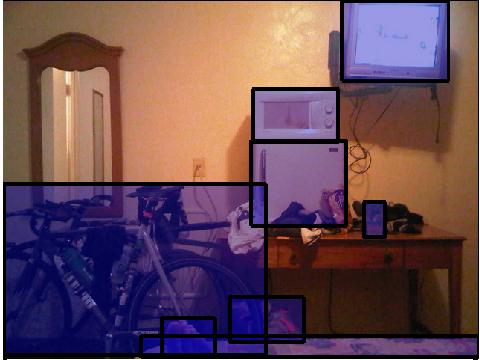}}\\
\bmvaHangBox{\includegraphics[width=3cm]{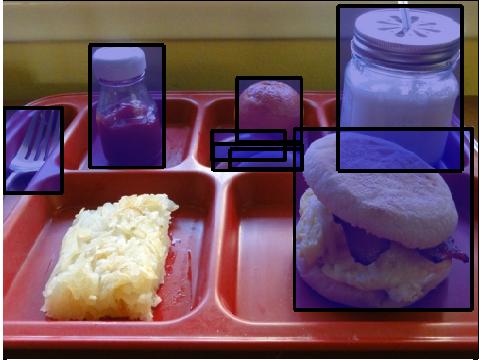}
\includegraphics[width=3cm]{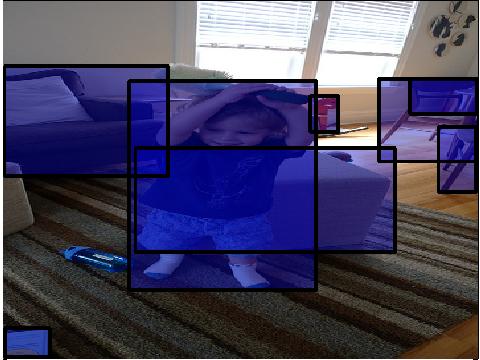}
\includegraphics[width=3cm]{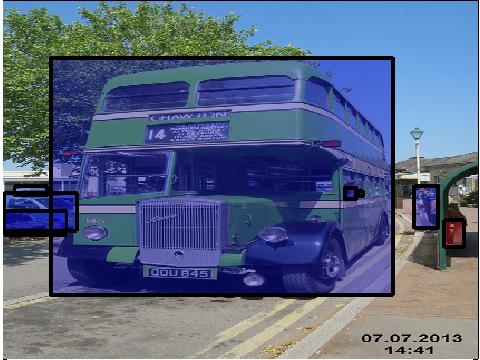}
\includegraphics[width=3cm]{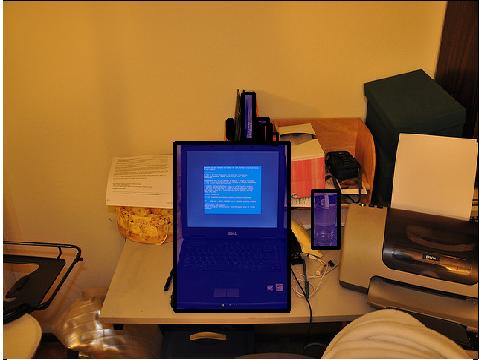}}\\
\bmvaHangBox{\includegraphics[width=3cm]{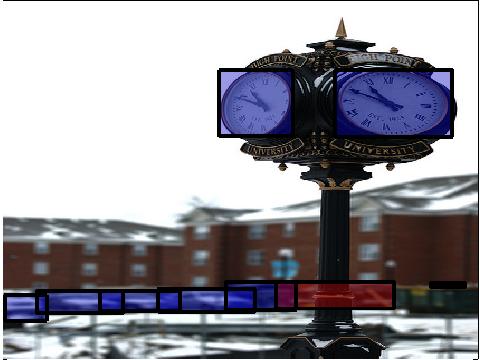}
\includegraphics[width=3cm]{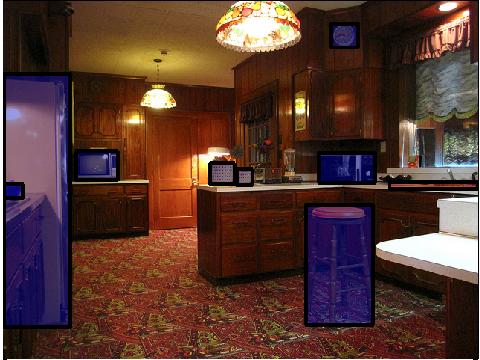}
\includegraphics[width=3cm]{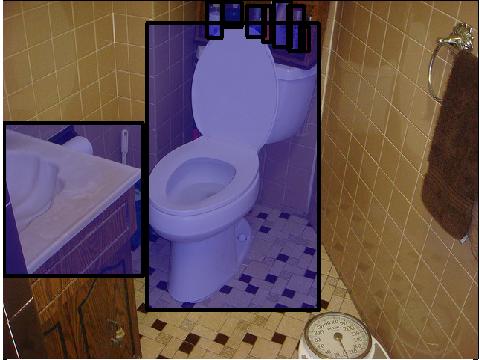}
\includegraphics[width=3cm]{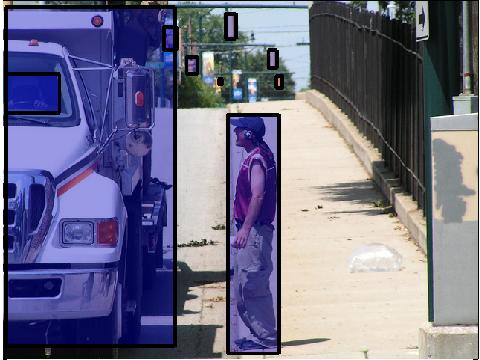}}\\
\bmvaHangBox{\includegraphics[width=3cm]{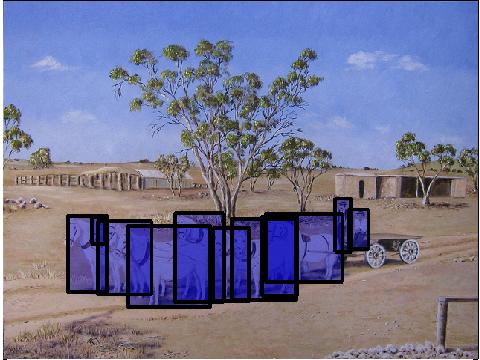}
\includegraphics[width=3cm]{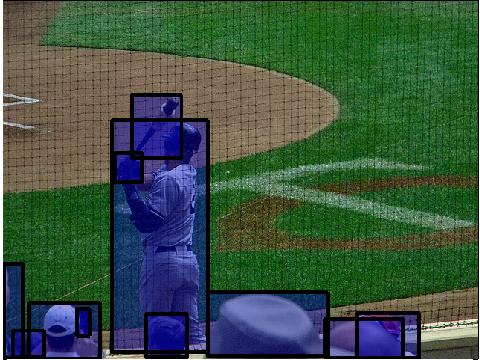}
\includegraphics[width=3cm]{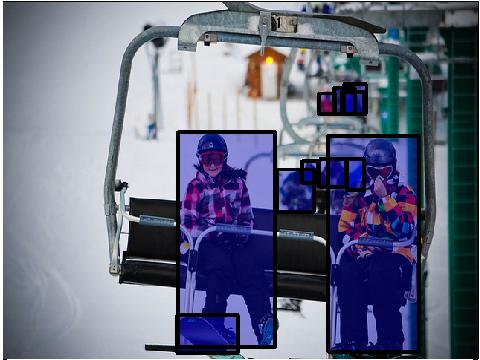}
\includegraphics[width=3cm]{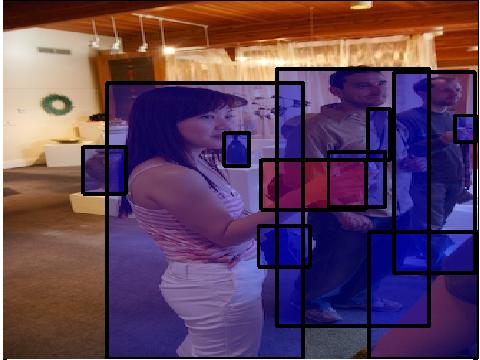}}\\
\bmvaHangBox{\includegraphics[width=3cm]{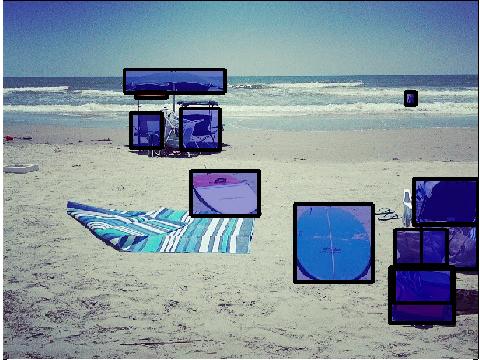}
\includegraphics[width=3cm]{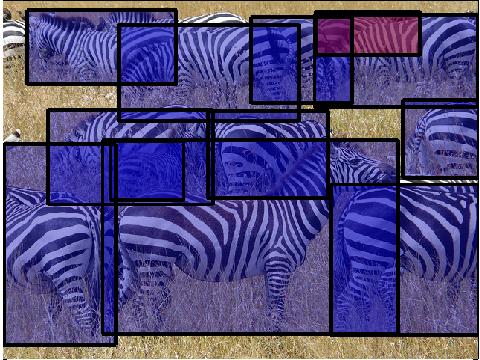}
\includegraphics[width=3cm]{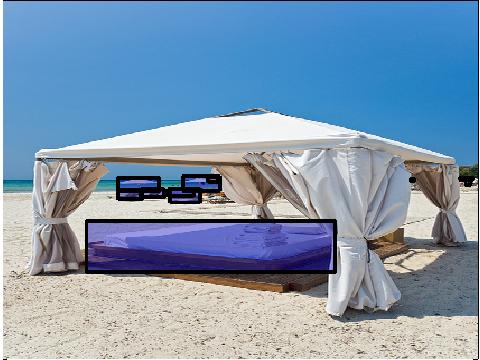}
\includegraphics[width=3cm]{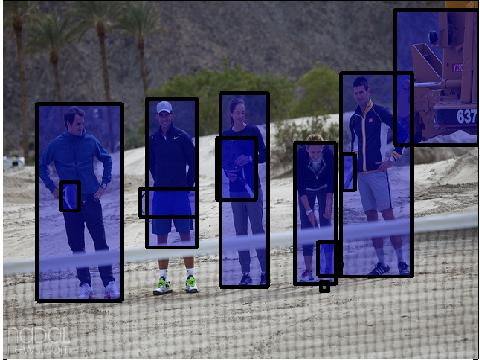}}\\
\bmvaHangBox{\includegraphics[width=3cm]{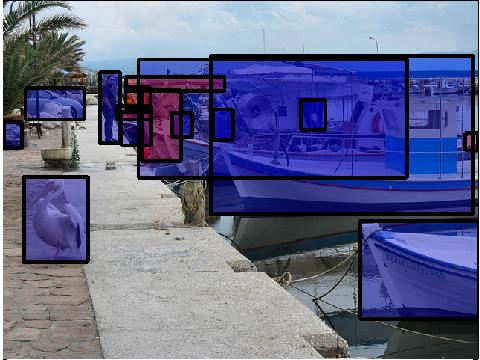}
\includegraphics[width=3cm]{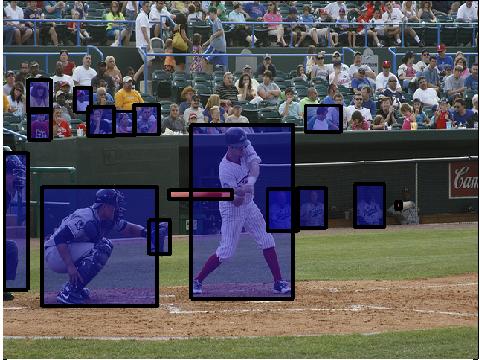}
\includegraphics[width=3cm]{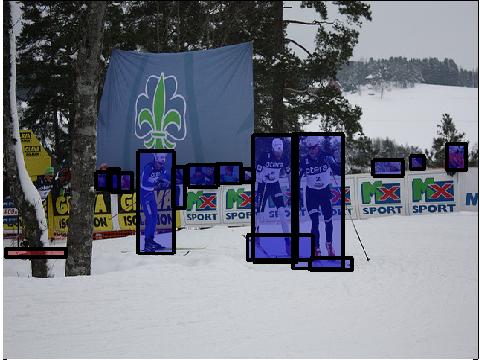}
\includegraphics[width=3cm]{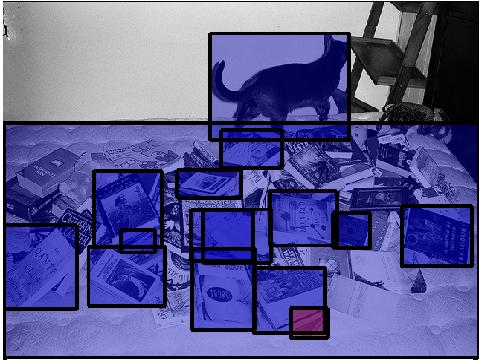}}\\
\bmvaHangBox{\includegraphics[width=3cm]{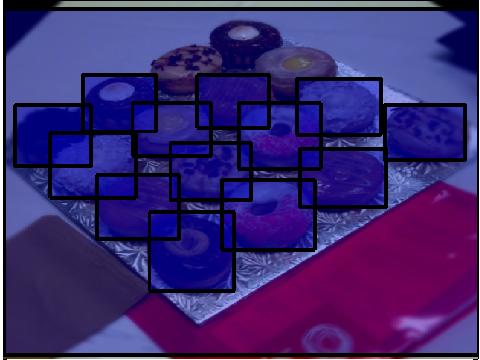}
\includegraphics[width=3cm]{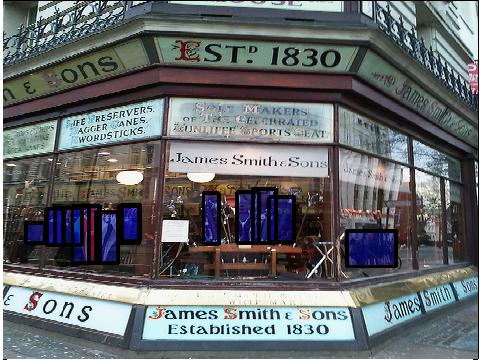}
\includegraphics[width=3cm]{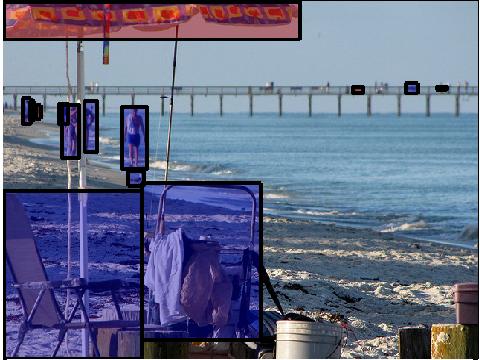}
\includegraphics[width=3cm]{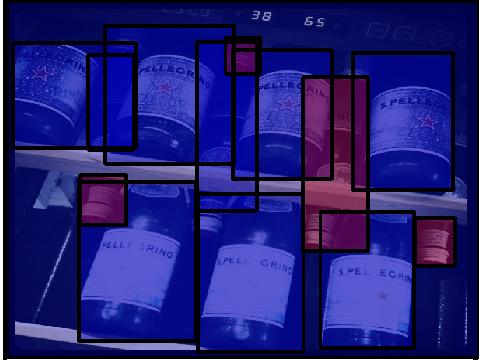}}\\
\end{center}
\vspace{-10pt}
\caption{
\small{\textbf{Qualitative results in COCO.} 
The blue rectangles are the box proposals generated by our approach that best localize (in terms of IoU) the ground truth boxes.
The red rectangles are the ground truth bounding boxes that were not discovered by our box proposal approach (their IoU with any box proposal is less than 0.5). Note that not all the object instances on the images are annotated.}}
\label{fig:qualitative}
\vspace{-10pt}
\end{figure}

In Figure~\ref{fig:qualitative} we provide qualitative results of our \emph{AttractioNet} box proposal approach on images coming from the COCO validation set. 
Note that our approach manages to recall most of the objects in an image, even in the case that the depicted scene is crowded with multiple objects that heavily overlap with each other.

\section{Conclusions} \label{sec:conclusions}
\vspace{-5pt}
In our work we propose a bounding box proposals generation method, which we call \emph{AttractioNet}, whose key elements are
a strategy for actively searching of bounding boxes in the promising image areas and
a powerful object location refinement module that extends the recently introduced LocNet~\cite{gidaris2016locnet} model
on localizing objects agnostic to their category. 
We extensively evaluate our method on several image datasets
(i.e. COCO, PASCAL, ImageNet detection and NYU-Depth V2 datasets)
demonstrating in all cases average recall results that surpass the previous state-of-the-art by a significant margin while also providing strong empirical evidence about the generalization ability of our approach w.r.t. unseen categories.
Even more,
we show the significance of our \emph{AttractioNet} approach in the object detection task 
by coupling it with a VGG16-Net based detector and thus managing to surpass the detection performance of all other VGG16-Net based detectors while even being on par with a heavily tuned ResNet-101 based detector. 
We note that, apart from object detection, there exist several other vision tasks, 
such as exemplar 2D-3D detection~\cite{MassaCVPR16}, visual semantic role labelling~\cite{gupta2015visual}, caption generation~\cite{karpathy2015deep} or visual question answering~\cite{shih2015look}, 
for which a box proposal generation step can be employed.
We are thus confident that our \emph{AttractioNet} approach could have a  significant value with respect to many other  important applications as well.

\section*{Acknowledgements}
This work was supported by the \emph{ANR SEMAPOLIS} project. 
We would like to thank Pedro O. Pinheiro for help with the experimental results and Sergey Zagoruyko for helpful discussions. 
We would also like to thank the authors of SharpMask~\cite{PinheiroLCD16} (Pedro O. Pinheiro, Tsung-Yi Lin, Ronan Collobert and Piotr Dollar) for providing us with its box proposals.

\bibliography{my_paper_bib_short}

\appendix

\section{Multi-threshold non-max-suppresion re-ordering} \label{sec:nms}

As already described in section~\ref{sec:algorithm}, at the end of our active box proposal generation strategy we include a non-maximum-suppression~\cite{felzenszwalb2010object} (NMS) step that is applied on the set $\textbf{C}$ of scored candidate box proposals in order then to take the final top $K$ output box proposals (see algorithm~\ref{algo:prop_gen}).
However, the optimal IoU threshold (in terms of the achieved AR) for the NMS step depends on the desired number $K$ of output box-proposals.
For example, for 10, 100, 1000 and 2000 proposals the optimal IoU thresholds are $0.55$, $0.75$, $0.90$ and $0.95$ respectively.
Since our plan is to make  our box proposal system publicly available, we would like to make its use easier for the end user. 
For that purpose, we first apply on the set $\textbf{C}$ of scored candidate box proposals a simple NMS step with IoU threshold equal to $0.95$ in order to then get the top 2000 box proposals and then we follow a multi-threshold non-max-suppression technique that re-orders this set of 2000 box proposals such that for any given number $K$ the top $K$ box proposals in the set better cover (in terms of achieved AR) the objects in the image.

Specifically, assume that $\{t_i\}_{i=1}^{N_k}$ are the optimal IoU thresholds for $N_k$ different desired numbers of output box proposals $\{K_i\}_{i=1}^{N_k}$, where both the thresholds and the desired number of box proposals are in ascending order\footnote{We used the IoU thresholds of $\{0.55, 0.60, 0.65, 0.75, 0.80, 0.85, 0.90, 0.95\}$ for the desired numbers of output box proposals $\{10, 20, 40, 100, 200, 400, 1000, 2000\}$. Those IoU thresholds were cross-validated on a validation set different than this used for the evaluation of our approach.}. 
Our multi-threshold NMS strategy starts by applying on the aforementioned set $\textbf{L}$ of 2000 box proposals simple single-threshold NMS steps with IoU thresholds $\{t_i\}_{i=1}^{N_k}$ that results on $N_k$ different lists of box proposals $\{\textbf{L}(t_i)\}_{i=1}^{N_k}$ (note that all the NMS steps are applied on the same list $\textbf{L}$ and not in consecutive order). 
Then, starting from the lowest threshold $t_1$ (which also is the more restrictive one) we take from the list $\textbf{L}(t_1)$ the top $K_{1}$ box proposals and we add them to the set of output box proposals $\textbf{P}$. 
For the next threshold $t_2$ we get the top $K_2 - |\textbf{P}|$ box proposals from the set $\{\textbf{L}({t_2}) \setminus \textbf{P}\}$ and again add them on the set $\textbf{P}$. 
This process continues till the last threshold $t_{N_k}$ at which point the size of the output box proposals set $\textbf{P}$ is $K_{N_k} = 2000$.
Each time $i=1,...,N_k$ we add box proposals on the set $\textbf{P}$, their objectness scores are altered according to the formula $\tilde{o} = o + (N_k-i)$ (where $o$ and $\tilde{o}$ are the initial and after re-ordering objectness scores correspondingly) such that their new objectness scores to correspond to the order at which they are placed in the set $\textbf{P}$. 
Note that this technique does not guarantee an optimal re-ordering of the boxes (in terms of AR),  however it works  sufficiently well in practice.

\section{Detection system} \label{sec:detection_system}

In this section we provide further implementation details about the object detection system used in~\S\ref{sec:detection}.\\

\textbf{Architecture.}
Our box proposal based object detection network consists of a Fast-RCNN~\cite{girshick2015fast} category-specific recognition module and a LocNet Combined ML~\cite{gidaris2016locnet} category-specific bounding box refinement module that share the same image-wise convolutional layers (conv1\_1 till conv5\_3 layers of VGG16-Net).\\

\textbf{Training.}
The detection network is trained on the union of the COCO train set that includes around $80k$ images and on a subset of the COCO validation set that includes around $35k$ images (the remaining $5k$ images of COCO validation set are being used for evaluation). 
For training we use our \emph{AttractioNet} box proposals and we define as positives those that have IoU overlap with any ground truth bounding box at least 0.5 and as negatives the remaining proposals. 
For training we use SGD where each mini-batch consists of 4 images with 64 box proposals each (256 boxes per mini-batch in total) and the ratio of negative-positive boxes is 3:1.
We train the detection network for $500k$ SGD iterations starting with a learning rate of $0.001$ and dropping it to $0.0001$ after $320k$ iterations.
We use the same scale and aspect ratio jittering technique that is used on \emph{AttractioNet} and is described in section~\ref{sec:training}.

\section{Common categories between ImageNet and COCO} \label{sec:imagenet_coco}

In this section we list the ImageNet detection task object categories that we identified to be present also in the COCO dataset. Those are:\\
{\sffamily
airplane, apple, backpack, baseball, banana, bear, bench,
bicycle, bird, bowl, bus, car, chair, cattle, 
computer keyboard, computer mouse, cup or mug, dog,
domestic cat, digital clock, elephant, horse, hotdog, laptop,
microwave, motorcycle, orange, person, pizza, refrigerator, sheep, ski,
tie, toaster, traffic light, train, zebra, racket, remote control, sofa, 
tv or monitor, table, watercraft, washer,  water bottle,
wine bottle, ladle, flower pot, purse, stove, koala bear,
volleyball, hair dryer, soccer ball, rugby ball, croquet ball,
basketball, golf ball, ping-pong ball, tennis ball}.

\section{Ignored NUY Depth dataset categories} \label{sec:ignored_categories}

In this section we list the 12 most frequent non-object categories that we identified on the NUY Depth V2 dataset:\\
{\sffamily
curtain, cabinet, wall, floor, ceiling, room divider, window shelf, stair, counter, window, pipe and column.}

\end{document}